\documentclass[twocolumn, final]{svjour3}   

\makeatletter
\expandafter\let\csname opt@amsmath.sty\endcsname\relax
\AtBeginDocument{
 \mathindent=15pt 
 \@mathmargin\@centering} 
\makeatother

\RequirePackage{fix-cm}
\usepackage{graphicx}
\usepackage{amsmath, amsfonts, amssymb, lipsum}
\usepackage{algorithm}
\usepackage{algorithmic}
\usepackage[misc]{ifsym}
\usepackage{color}
\usepackage[nointegrals]{wasysym}
\usepackage{bm}
\usepackage{cite}
\usepackage{overpic}
\usepackage{appendix}
\usepackage{url}

\smartqed 

\numberwithin{equation}{section}
\newcommand\dd{\mathrm{d}}
\newcommand\pp{\partial}
\newcommand\tr{\mathrm{tr}}

\newcommand\KL{\mathrm{KL}}

\newcommand\x{\bm{x}}

\newcommand\uvec{\mathbf{u}}
\newcommand\X{\mathbf{X}}
\newcommand\y{\bm{y}}
\newcommand\z{\bm{z}}

\newcommand\F{{\sf F}}

\usepackage[normalem]{ulem}

\begin{document}

\title{Particle-based Energetic Variational Inference}

\author{
Yiwei Wang \and
Jiuhai Chen \and
Chun Liu \and
Lulu Kang
}

\institute{
Yiwei Wang \\
\email{ywang487@iit.edu}\\
Jiuhai Chen \\
\email{jchen168@hawk.iit.edu} \\
Chun Liu \\
\email{cliu124@iit.edu} 
Lulu Kang \Letter \\
\email{lkang2@iit.edu} \\
Department of Applied Mathematics, Illinois Institute of Technology, IL, USA
}


\maketitle

\begin{abstract}
We introduce a new variational inference (VI) framework, called \emph{energetic variational inference} (EVI).
It minimizes the VI objective function based on a prescribed \emph{energy-dissipation law}.
Using the EVI framework, we can derive many existing Particle-based Variational Inference (ParVI) methods, including the popular Stein Variational Gradient Descent (SVGD) approach.
More importantly, many new ParVI schemes can be created under this framework.
For illustration, we propose a new particle-based EVI scheme, which performs the particle-based approximation of the density first and then uses the approximated density in the variational procedure, or ``Approximation-then-Variation'' for short. 
Thanks to this order of approximation and variation, the new scheme can maintain the variational structure at the particle level, and can significantly decrease the KL-divergence in each iteration.  
Numerical experiments show the proposed method outperforms some existing ParVI methods in terms of fidelity to the target distribution.

\noindent{\bf Keywords}: KL-divergence; Energetic variational approach; Gaussian mixture model; Kernel function; Implicit-Euler; Variational inference.
\end{abstract}

\section{Introduction}

Bayesian methods play an important role in statistics and data science nowadays. 
They provide a rigorous framework for uncertainty quantification of various statistical learning models \cite{stuart2010inverse, gelman2013bayesian}.
The main components of a Bayesian model includes a set of observational data $\{\y_i\}_{i=1}^I$ with $\y_i\in\mathbb{R}^D$, the model assumption of the likelihood $\rho(\{\y_i\}_{i=1}^I|\x)$ with certain unknown parameters $\x \in \mathbb{R}^d$, and a user-specified prior distribution for the parameters $\rho_0(\x)$.
The key step in Bayesian inference is to obtain the posterior distribution, denoted by $\rho(\x | \{ \y_i \}_{i=1}^I)$.
Following the Bayes' theorem, the posterior distribution of the unknown parameters is 
\[\rho (\bm{\x}|\{\y_i\}_{i=1}^I)=\frac{\rho(\{\y_i\}_{i=1}^I|\bm{\x})\rho_0(\bm{\x})}{\rho( \{\y_i\}_{i=1}^I)}.\]
However, it is a long standing challenge to obtain the posterior distribution in practice when the analytical formula of $\rho (\bm{\x}|\{\y_i\}_{i=1}^I)$ is not tractable due to the integration $\rho( \{\y_i\}_{i=1}^I)=\int \rho(\{\y_i\}_{i=1}^I|\bm{\x})\rho_0(\bm{\x}) \dd \x$.

Many approximate inference methods have been developed to approximate the posterior distribution. 
Among them, two popular classes of methods are Markov Chain Monte Carlo (MCMC) algorithms \cite{metropolis1953equation, hastings1970monte, geman1984stochastic, welling2011bayesian} and Variational Inference (VI) methods \cite{jordan1999introduction, neal1998view, wainwright2008graphical, blei2017variational}.
MCMC is a family of methods that generate samples by constructing a Markov chain whose equilibrium distribution is the target distribution. 
Examples include the Metropolis--Hastings algorithm \cite{metropolis1953equation, hastings1970monte}, Gibbs sampling \cite{geman1984stochastic, casella1992explaining}, Langevin Monte Carlo (LMC) \cite{rossky1978brownian, parisi1981correlation, roberts1996exponential, welling2011bayesian}, and Hamiltonian Monte Carlo (HMC) \cite{neal1993probabilistic, duane1987hybrid}. 

The VI framework essentially transforms the inference problem into an optimization problem, which minimizes some kind of objective functional over a prescribed family of distributions denoted by $\mathcal{Q}$ \cite{blei2017variational}. 
The objective functional measures the difference between a candidate distribution in $\mathcal{Q}$ and the target distribution. 
For Bayesian models, the target distribution is the posterior distribution. 
VI has a wide application that goes beyond Bayesian statistics and is a powerful tool for approximating probability densities.
A common choice of the objective functional is the Kullback-Leibler (KL) divergence \cite{blei2017variational}. 
For any two distributions $\rho(\x)$ and $\rho^*(\x)$, the KL-divergence from $\rho$ to $\rho^*$ is given by 
\begin{equation}\label{eq:KL}
\KL( \rho(\x) ||\rho^*(\x)) = \int \rho(\x) \ln \left( \frac{\rho(\x)}{\rho^*(\x )} \right) \dd \x. 
\end{equation}

In the review paper \cite{blei2017variational}, the authors have made a detailed comparison between MCMC and VI methods, both conceptually and numerically. 
Essentially, MCMC methods guarantee the convergence of the generated samples to the target distribution when certain conditions are met. 
But the price of this asymptotic property is that MCMC methods tend to be more computationally intensive and thus might not be suitable for large datasets. 
On the contrary, VI methods do not have this asymptotic guarantee.
For particle-based VI, the fidelity of the empirical distribution of the particles to the target distribution depends on the VI algorithm as well as the number of particles. 
On the other hand, since VI is essentially an optimization problem and it can take advantage of the stochastic optimization methods, VI methods can be significantly faster than MCMC. 
For detailed differences and connections between the two types of methods, see \cite{mackay2003information} and \cite{salimans2015markov}. 
In this paper, we only focus on the VI approaches.

The VI framework minimizes $\KL( \rho(\x) ||\rho^*(\x))$ with respect to $\rho \in \mathcal{Q}$ in order to approximate the target distribution $\rho^*$. 
In traditional VI methods \cite{blei2017variational}, $\mathcal{Q}$ is often taken as a family of parametric distributions. 
There also have been growing interests in flow-based VI methods, in which $\mathcal{Q}$ consists of distributions obtained by a series of smooth transformations from a tractable initial reference distribution. 
Examples include normalizing flow VI methods \cite{rezende2015variational, kingma2016improved, salman2018deep} and particle-based VI methods (ParVIs) \cite{liu2016stein, liu2017stein, liu2018riemannian, chen2018unified, liu2019understanding, chen2019projected}. 
One ParVI method that has attracted much attention is the Stein Variational Gradient Descent (SVGD) \cite{liu2016stein, detommaso2018stein, wang2019stein, li2019stochastic}. 
Many existing ParVI methods can be viewed as some version of the approximated Wasserstein gradient flow of the KL-divergence \cite{liu2017stein}. 
As explained in Section 3, these methods may not preserve the variational structure at the particle level because approximation of the density function is performed after the variational step.

In this paper, we introduce a new variational inference framework, named as \emph{Energetic Variational Inference} (EVI). 
It consists of two ingredients, a continuous formulation of the variational inference, and a discretization strategy that leads to a practical algorithm.
Inspired by the non-equilibrium thermodynamics, we propose using a \emph{energy-dissipation law} to describe the mechanism of minimizing the VI objective functional, for instance, the KL-divergence. 
An energetic variational inference algorithm can be obtained by employing an energetic variational approach and a proper discretization.
Using the EVI framework, we can derive and explain many existing ParVI methods, such as the SVGD method. 
More importantly, many new ParVI schemes can be created under the EVI framework. 
To demonstrate how to develop a new EVI method, we propose a new particle-based EVI scheme, which performs the particle-based approximation of the density first and then uses the approximated density in the variational procedure. 
Thanks to this ``Approximation-then-Variation" order, we can derive a system of ordinary differential equations (ODEs) of particles that preserves the variational structure at the particle level, which is different from many existing methods. 
Such an ODE system can be solved via the implicit Euler method, which can be reformulated into an optimization problem. 
By the virtue of the variational structure at the particle level, we can significantly decrease the discretized KL-divergence in every iteration and push the density of the particles close to the target distribution efficiently.

In the remaining sections, we first introduce some preliminary background on the flow map and the energetic variational approach that is commonly used in mathematical modeling in Section 2. 
In Section 3, we propose the energetic variational inference (EVI) framework. 
Specifically, we first lay out the general continuous formulation of the EVI, and then introduce two different ways to discretize the continuous EVI.
One is the ``Approximation-then-Variation'' approach and the other is the ``Variation-then-Approximation'' approach. 
Both lead to particle-based EVI methods.
The dynamics of the particles are described by an ODE system, which can be solved by explicit or implicit Euler methods. 
Using the implicit-Euler, we propose one new example of particle-based EVI, called EVI-Im. 
In Section 4, we compare the EVI-Im with some existing particle-based VI methods. 
The paper is concluded in Section 5.

\section{Preliminary}

Before reviewing the preliminary topics on flow map and energetic variational approaches, we first clarify some notations used in this paper. 
Let $f(\bm x, t)$ be a scalar function of $d$-dimensional space variable $\bm x\in \mathcal{X} \subseteq \mathbb{R}^d$ and time $t \in [0, \infty)$. 
We denote the derivative of $f(\bm x, t)$ with respect to $t$ as $\dot{f}(\bm x, t)$ or $\dot{f}$ for short, and thus $\dot{f}$ is still a scalar function.  
The gradient of $f(\bm x, t)$ with respect to $\bm x$ is $\nabla f(\x, t)$ or $\nabla f$, and thus $\nabla f$ is a $d$-dimensional function. 
When the time $t$ is taken at a series of discrete values, i.e., $t=0, 1, 2,\ldots$, the discrete $t$ becomes the index of a sequence of function values of $f$ evaluated at $\x$. 
We write the integer index in the superscript position of $f$, i.e., $f^t(\x)$. 
The subscript position of $f$ is to label different functions. 

\subsection{Minimizing KL-Divergence Through Flow Maps}

The goal of the variational inference is to find a density function $\rho$ from a family of density functions $\mathcal{Q}$ by minimizing the VI objective functional, such as the KL-divergence from $\rho(\bm x)$ to the target density function $\rho^*(\bm x)$. 
The complexity of this optimization problem is decided by the feasible region, i.e., the family $\mathcal{Q}$. 
Traditional variational inference methods choose $\mathcal{Q}$ as a parametric family of probability distributions. 
For example, the mean-field variational family assumes the mutual independence between the $d$ dimensions of random variable $\bm x$, i.e., $\rho(\bm x)=\prod_{i=1}^d \rho_i(x_i)$, where $\rho_i$ is a density function from a user specified family of one-dimensional probability densities \cite{bishop2006pattern, blei2017variational}.

In the flow-based VI methods, the set $\mathcal{Q}$ consists of distributions obtained by smooth transformations of a tractable initial reference distribution \cite{li2019stochastic}.
The idea of using maps to transform a distribution to another has been explored in many earlier papers \cite{tabak2010density, el2012bayesian}. 
Specifically, given a tractable reference distribution $\rho_0(\bm z): \mathcal{X}^0\rightarrow \mathbb{R}^+$ and a sufficiently smooth one-to-one map $\bm \phi(\cdot)$, such that $\bm x=\bm \phi(\bm z)$, the family $\mathcal{Q}$ is defined by
\begin{equation}\label{eq:Q}  
\begin{aligned} 
&\mathcal{Q}= \{\rho_{[\bm \phi]}(\bm x) = \rho_0(\bm \phi^{-1}(\bm x))\left\vert\det[\nabla_{\x} \bm \phi^{-1}(\x)]\right\vert, \x=\bm \phi(\z), \\
& \bm \phi: \mathcal{X}^0\rightarrow \mathcal{X}\text{ is a smooth one-to-one map.}\}
\end{aligned}
\end{equation}
We assume $\mathcal{X}^0 = \mathcal{X} = \mathbb{R}^d$ throughout this paper, but all the results can be generalized to the case where $\mathcal{X}^0 \neq \mathcal{X}$. 
Moreover, since $\bm \phi$ is one-to-one, we can enforce $\det[\nabla_{\z} \bm{\phi}(\z)]> 0$.

Given $\mathcal{Q}$ in \eqref{eq:Q}, solving the following problem 
\begin{equation}\label{eq:VI}
\rho_{\text{opt}} =\text{arg}\min_{\rho \in \mathcal{Q}}\text{KL}(\rho||\rho^*)
\end{equation}
is equivalent to finding the optimal smooth one-to-one map $\bm \phi_{\text{opt}}$ such that 
$$\rho_{\text{opt}}(\x)=\rho_0(\bm \phi^{-1}_{\text{opt}}(\bm x))\det[\nabla_{\x} \bm \phi_{\text{opt}}^{-1}(\x)].$$
As in many optimization approaches, we expect it requires a number of transformations, say $K$ steps, to find the optimal map, or equivalently, 
$$
\bm \phi_{\text{opt}}(\cdot)=\bm \psi^K\circ \bm \psi^{K-1}\ldots \circ \bm \psi^1 (\cdot).
$$
Each $\bm \psi^t(\cdot)$ is a smooth and one-to-one map such that $\bm x^t=\bm \psi^t(\bm x^{t-1})$. 
At the $t$th step, suppose $\bm \phi^t(\cdot) = \bm \psi^{t} \circ \bm \psi^{t-1}\ldots \circ \bm \psi^1 (\cdot)$ is a proper transform, then
$$
\begin{aligned}
\rho^t(\x^t) & =\rho^{t-1}((\bm \psi^t)^{-1}(\x^{t})) \det[\nabla (\bm \psi^t)^{-1}(\x^{t})] \\
 & = \rho_{0}((\bm \phi^t)^{-1}(\x^t)) \det[\nabla (\bm \phi^t)^{-1}(\x^t)].
\end{aligned}
$$

Intuitively, the series of transformations should move the initial density $\rho_0$ closer and closer to the target density $\rho^*$ and eventually achieve convergence in terms of the KL-divergence. 
Therefore, $\text{KL}(\rho^t || \rho^{*})$ should be decreased after each step, i.e., $$\text{KL}(\rho^t || \rho^{*})-\text{KL}(\rho^{t-1}|| \rho^{*})\leq 0.$$
If we generalize the meaning of $t$ from the discrete step index to the continuous time $t \in[0, \infty)$, we can consider $\rho^t(\bm x)$ as a density function evolving continuously with respect to time $t$. 
To emphasize this point, we use the notation $\rho(\bm x,t)$ instead of $\rho^t(\bm x)$. 
Therefore, $\KL(\rho(\bm x,t)||\rho^*(\x))$ should be decreased with respect to $t$, i.e., 
$$\frac{\dd}{\dd t} \KL(\rho(\bm x,t)||\rho^*(\x)) \leq 0.$$
The key to minimizing the KL-divergence is to determine the speed of decreasing $\KL(\rho(\bm x,t)||\rho^*(\x))$. 
In Section 3, we show how to use the energy-dissipating law to specify $\frac{\dd}{\dd t} \KL(\rho(\x,t)||\rho^*(\x))$.

When $t$ is generalized to continuous time, $\bm \phi^t(\cdot)$ becomes a smooth one-to-one map that also continuously evolves. 
Therefore, we use the notation $\bm \phi(\cdot, t)$ instead of $\bm \phi^t(\cdot)$. 
Since $\bm \phi(\cdot, t)$ is a smooth one-to-one map, it can be defined through a smooth, bounded velocity field $\uvec \in \mathbb{R}^d \times [0, \infty)$ as in Definition \ref{def:vfield}. 
This definition is also used in \cite{sonoda2019transport}. 
\begin{definition}\label{def:vfield}
Given a smooth and bounded velocity field $\uvec: \mathbb{R}^d \times [0, \infty) \rightarrow \mathbb{R}^d$, a flow map $\bm \phi(\z, t): \mathbb{R}^d\times [0, \infty) \rightarrow \mathbb{R}^d$ is a map specified by an ordinary differential equation (for any fixed $\z$)
\begin{equation}\label{eq:phi}
\begin{cases}
& \dot{\bm \phi}(\z, t)= \uvec(\bm \phi(\z, t), t), \quad \z \in \mathbb{R}^d, \quad t > 0 \\
& \bm \phi(\z,0) = \z, \quad \z \in \mathbb{R}^d. \\
\end{cases} 
\end{equation}
\end{definition}

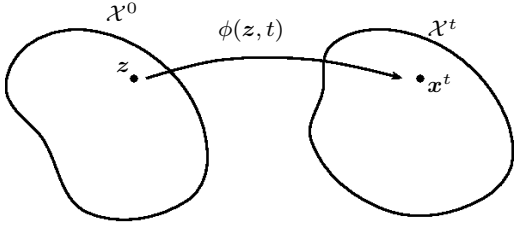
\begin{figure}[ht]
\centering
\begin{overpic}[width = 0.8 \linewidth]{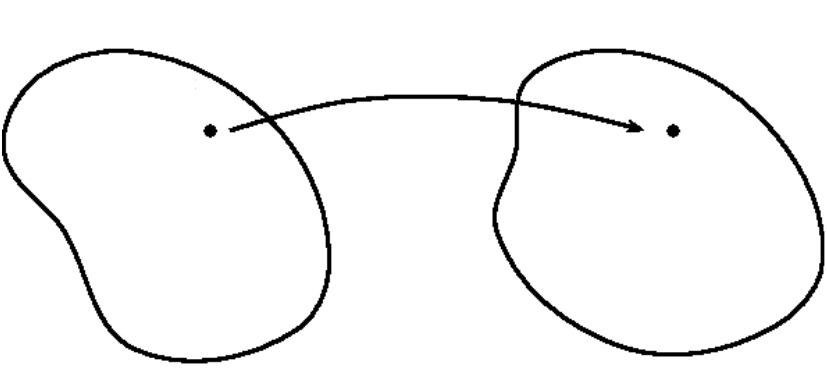}
\put(20, 40){$\mathcal{X}^0$}
\put(22, 30){$\z$}
\put(42, 37){$\phi(\z, t)$}
\put(83, 26){$\x^t$}
\put(83, 37){$\mathcal{X}^t$}
\end{overpic}
\caption{An illustration of a flow map $\bm \phi(\z, t)$.\label{FM}}
\end{figure}

 In continuum mechanics, $\bm \phi(\z, t)$ is known as the \emph{flow map} \cite{temam2005mathematical, gonzalez2008first}.
To better illustrate the idea of the flow map, we plot it conceptually in Fig. \ref{FM}. 
For fixed $\z$, $\bm \phi(\z, t)$ is the trajectory of a particle (or sample) with initial position $\z$. 
For fixed $t$, $\bm \phi(\z, t)$ is a diffeomorphism between $\mathcal{X}^0$ (the initial domain) and $\mathcal{X}^t$ (the domain after $t$ transformations). 
An intuitive interpretation of $\uvec$ is that $\uvec$ is the speed of the probability mass, which is transported due to the transformation $\bm \phi(\cdot, t)$.
But directly finding $\bm \phi(\cdot, t)$ is a difficult task.
Thanks to this relationship \eqref{eq:phi}, we can decide the transform $\bm \phi(\cdot, t)$ by specifying $\uvec$.

When the flow map is defined by the velocity field $\uvec(\x, t)$, the corresponding distribution $\rho(\x, t)$ is given by 
\[\rho(\bm \phi(\z, t), t) = \frac{\rho_0(\z)}{\det[\nabla_{\z} \bm{\phi}(\z, t)]}.\]
The relation between $\uvec$ and $\rho(\x, t)$ is described by the following proposition. 
\begin{proposition}\label{thm:transport} {\bf (Transportation equation)}
If $\bm \phi(\z, t)$ satisfies \eqref{eq:phi} and $\x=\bm \phi(\z, t)$, the time-dependent probability density $\rho(\x, t)$, which is induced by $\bm \phi(\z,t)$, satisfies the transport equation
\begin{equation}\label{Transport}
\begin{cases}
& \dot{\rho} + \nabla \cdot (\rho \uvec) = 0, \\
& \rho(\x, 0) = \rho_0(\x),
\end{cases}
\end{equation}
where $\rho_0$ is the initial density and $\dot{\rho}$ is the derivative of $\rho(\x,t)$ with respect to $t$.
\end{proposition}
Equation \eqref{Transport} is known as the transport equation or continuity equation \cite{villani2008optimal, sonoda2019transport}. 
Its derivation is in Appendix B. 
Definition \ref{def:vfield} and Proposition \ref{thm:transport} indicate that we only need to determine the transport velocity $\uvec\left(\x, t \right)$ so as to determine the flow map $\bm \phi(\z, t)$ and $\rho({\bm \phi}(\z, t), t)$.  

\subsection{Energetic Variational Approach}
Here we briefly introduce the energetic variational approach in mathematical modeling \cite{liu2009introduction, giga2017variational}, which is originated from the pioneering work of L. Rayleigh \cite{rayleigh1873note} and Onsager \cite{onsager1931reciprocal, onsager1931reciprocal2}. 
It provides a unique way to determine the dynamics of a system via a prescribed \emph{energy-dissipation law}
\begin{equation}\label{EDL1}
\frac{\dd}{\dd t} \mathcal{F}[\bm \phi] = - 2 \mathcal{D} [{\bm \phi}, \dot{\bm \phi}],
\end{equation}
which describes how the total energy of the system decreases with respect to time.
Here $\mathcal{F}$ is the Helmholtz free energy, $-2 \mathcal{D} \leq 0$ is the rate of energy dissipation, $\bm{\phi}$ is the state variable of the system, and $\dot{\bm \phi}$ is the derivative of $\bm \phi$ with respect to $t$. 
For a given energy-dissipation law (\ref{EDL1}), the energetic variational approach derives the dynamics of the system (or how energy $\mathcal{F}$ dissipates over time) through two variational procedure, the Least Action Principle (LAP) and the Maximum Dissipation Principle (MDP), which leads to  
\begin{equation}\label{eq_fb}
\frac{\delta \mathcal{D}}{\delta \dot{\bm\phi}} = - \frac{\delta \mathcal{F}}{\delta \bm{\phi}}, 
\end{equation}
where $\frac{\delta \mathcal{F}}{\delta {\bm \phi}}$ denotes the Fr\'echet derivative of $\mathcal{F}$ with respect to ${\bm \phi}$, defined as
$(\frac{\delta \mathcal{F}}{\delta {\bm \phi}}, {\bm \psi}) = \lim_{\epsilon \rightarrow 0} \frac{\mathcal{F}({\bm \phi} + \epsilon {\bm \psi}) - \mathcal{F}({\bm \phi})}{\epsilon}$, and $\frac{\delta \mathcal{D}}{\delta \dot{\bm \phi}}$ denotes the Fr\'echet derivative of $\mathcal{D}$ with respect to $\dot{\bm \phi}$.
More details on the energetic variational approach and the derivation of \eqref{eq_fb} are shown in Appendix A. 

\section{Energetic Variational Inference}

\subsection{Continuous Formulation}
In this subsection, we first propose a continuous formulation of EVI. 
The idea is to specify the dynamics of minimizing KL-divergence via an \emph{energy-dissipation law}, and we can employ the energetic variational approach to obtain the equation of the flow map ${\bm \phi(\z, t)}$. 
More specifically, as an analogy to physics, the KL-divergence is viewed as the \emph{Helmholtz free energy}\cite{murphy2012machine}, i.e., $\mathcal{F}[{\bm \phi}] = \KL(\rho(\x, t) || \rho^*)$. 
The free energy $\mathcal{F}$ depends on $\bm \phi$ since $\x=\bm \phi(\z, t)$ as a result of the flow map. 
We can impose an energy-dissipation law 
\begin{equation}\label{KL_D}
\frac{\dd}{\dd t} \KL(\rho(\x, t) || \rho^{*}) = - \int \eta(\rho) ||\dot{\bm \phi}||^2 \dd \x,
\end{equation}
where $\mathcal{D} =\frac{1}{2}\int \eta(\rho) ||\dot{\bm \phi}||^2 \dd \x$. 
Because the flow map $\bm \phi$ can be defined by the velocity field $\uvec$ (Definition \ref{def:vfield}), thus
\[\mathcal{D} =\frac{1}{2}\int \eta(\rho) ||\dot{\bm \phi}||^2 \dd \x=\frac{1}{2}\int \eta(\rho) ||\uvec||^2 \dd \x \geq 0.\]
The functional $\eta(\rho)$ is a user-specified functional of $\rho$ satisfying $\eta(\rho) > 0$ if $\rho > 0$. 
We denote $||\bm{a}|| = \sqrt{\bm{a}^{\rm T} \bm{a}}$ for $\bm a \in \mathbb{R}^d$ as the $l_2$ norm of a vector. 

Since $\rho(\x, t) = \rho_0(\bm{\phi}^{-1}(\x, t))$ is determined by $\bm{\phi}(\z, t)$ for a given $\rho_0(\z)$, the KL-divergence can be viewed as a functional of ${\bm \phi}$. 
By taking variation of $\KL$ with respect to ${\bm \phi}$ (see Appendix C for the detailed derivation), we can obtain
\begin{equation}\label{LAP}
\begin{aligned} 
-\frac{\delta \KL(\rho_{[\bm \phi]} || \rho^{*})}{\delta \bm \phi} = - (\nabla \rho + \rho \nabla V),
\end{aligned}
\end{equation}
where $\rho_{[\bm \phi]}(\x, t) = \rho_0(\bm{\phi}^{-1}(\x, t))$ and $V=-\ln \rho^*$.

Meanwhile, taking variational of $\mathcal{D}$ with respect $\dot{\bm \phi}$ yields
\begin{equation*}
\frac{\delta \mathcal{D}}{\delta \dot{\bm \phi}} = \eta(\rho)\dot{\bm \phi}.
\end{equation*}
Then according to \eqref{eq_fb}, $\dot{\bm \phi}$, i.e., the transport velocity $\uvec$ satisfies
\begin{equation}\label{velocity}
\eta(\rho) \dot{\bm \phi} = - ( \nabla \rho + \rho \nabla V).
\end{equation}
This equation gives us the specification of the transport velocity $\uvec$ based on the energy-dissipation law (\ref{KL_D}). 
Thanks to the transport equation \eqref{Transport}, $\rho$ can be obtained from the specified $\uvec$.
Therefore, \eqref{velocity} can be used to find the $\rho$ that minimizes the KL-divergence in the admissible set. 
Indeed, combining \eqref{velocity} with the transport equation \eqref{Transport}, we have 
\begin{equation}\label{CD_1}
\dot{\rho} = \nabla \cdot \left( \frac{\rho}{\eta(\rho)}( \nabla \rho + \rho \nabla V) \right),
\end{equation}
which is the continuous differential equation formulation for $\rho$. 
One can choose $\eta(\rho)$ to control the dynamics of the system. 
In the remainder of the paper, we choose $\eta(\rho) = \rho$, which is consistent with Wasserstein gradient flow \cite{jordan1998variational, santambrogio2017euclidean, frogner2018approximate}. 
We should emphasize the above derivation is rather formal. 
Under the suitable assumptions, one can show the existence of ${\bm \phi}(\z, t)$ for the equation \eqref{velocity}. 
We refer interested readers to \cite{evans2005diffeomorphisms, ambrosio2006stability, carrillo2010asymptotic} for theoretical discussions.

\begin{remark}
Different choices of dissipation laws lead to different dynamics to an equilibrium. 
For instance, we can take the energy-dissipation law in \cite{liu2020variational}
\begin{equation}
\label{eq:anotherlaw}
\frac{\dd}{\dd t} \KL(\rho(\x, t) || \rho^{*}) = - \int \left(\eta(\rho) ||\uvec||^2 + \nu(\rho)||\nabla \uvec||^2 \right)\dd \x.
\end{equation}
In this paper, we show that even with the simplest choice of the dissipation $\eta(\rho)=\rho$, the EVI framework can already lead to several new and existing ParVI methods.  
We will study the benefit of other choices of dissipation in future work.
\end{remark}

\subsection{Particle-based EVI}

In practice, there are two ways to approximate a probability density in $\mathcal{Q}$ defined in \eqref{eq:Q}. 
One is to approximate the transport map ${\bm \phi}(\z, t)$ directly, as used in variational inference with normalizing flow \cite{rezende2015variational}. 
The transport map can be approximated either by a family of parametric transformations \cite{rezende2015variational} or a piece-wise linear map \cite{carrillo2018lagrangian, liu2019lagrangian}. The main difficulty in such approaches is how to compute $\det[\nabla_{\z}\bm{\phi}(\z, t)]$ efficiently. 
We refer readers to \cite{rezende2015variational, carrillo2018lagrangian, liu2019lagrangian, papamakarios2019normalizing} for details.

Alternatively, a probability density in $\mathcal{Q}$ can be approximated by an empirical measure defined by a set of sample points $\{ \x_i(t) \}_{i=1}^N$. 
As used in many ParVI methods, 
\begin{equation}\label{emp}
\rho(\x, t) \approx \rho_{N}(\x, t) = \frac{1}{N} \sum_{i = 1}^N \delta (\x - \x_i(t)), 
\end{equation}
where $\x_i(t) = \bm{\phi}(\x_i(0), t)$ and $\x_i(0)$ is sampled from the initial reference distribution $\rho_0$. 
The sample points $\{\x_i(t)\}_{i=1}^N$ at time $t$ are called ``particles'' in the ParVIs literature. 
Instead of computing the map $\bm{\phi}(\z, t)$ explicitly at each time-step, only $\{\x_i(t)\}_{i=1}^N$ are computed in ParVIs. 
One can view this as a deterministic method to sample from the posterior.
The evolution of particles $\{\x_i(t)\}_{i=1}^N$ can be characterized by a system of ODEs, and it can be derived from the energy-dissipation law \eqref{KL_D} using the proposed EVI framework, as shown in the follows. 

There are two ways to derive such an ODE system. 
For short, we call them ``Approximation-then-Variation'' and ``Variation-then-Approximation'' approaches. 
Essentially, the two approaches use different orders of density approximation and variational procedure, which may lead to different ODE systems.

The {\bf Approximation-then-Variation} approach starts with a discrete energy-dissipation law
\begin{equation}
\frac{\dd}{\dd t} \mathcal{F}_h( \{ \x_i(t) \}_{i=1}^N) = - 2 \mathcal{D}_h( \{ \x_i(t) \}_{i=1}^N, \{ \dot{\x}_i(t) \}_{i=1}^N),
\end{equation}
which can be obtained by inserting the empirical approximation \eqref{emp} into the continuous energy-dissipation law with a suitable \emph{kernel regularization}. 
For instance, a discrete version of \eqref{KL_D}, which is the proposed dissipation mechanism of the KL-divergence, can be obtained by applying the particle approximation $\rho_N(\x, t)$ to \eqref{KL_D}. 
To avoid $\ln\delta(\x-\x_i(t))$ operation, we replace $\rho_N$ by the convolution $K_h*\rho_N$ inside the $\log$ function, where $K_h$ is a kernel function. 
This particle-based approximation leads to the regularized energy-dissipation law
\begin{equation}\label{Energy_Blob}
\frac{\dd}{\dd t} \int \rho_N\ln(K_h * \rho_N) + V \rho_N\dd \x = - \int_{\Omega} \rho_N||\uvec||^2 \dd \x,
\end{equation}
where
\begin{equation*}
K_h*\rho_N = \int K_h(\x - \y) \rho_N(\y, t) \dd y = \frac{1}{N} \sum_{j=1}^N K_h(\x - \x_j(t)).
\end{equation*}
We denote $K_h(\x - \x_j)$ by $K_h(\x, \x_j)$, which is a more conventional notation in the literature.
A typical choice of $K_h$ is the Gaussian kernel
\begin{equation}
K_h(\x_1,\x_2) = \frac{1}{(\sqrt{2\pi h})^d} \exp \left(- \frac{||\x_1-\x_2||^2}{h^2} \right).
\end{equation}
The regularized free energy \eqref{Energy_Blob} is proposed in \cite{carrillo2019blob} and has been used to design the Blob variational inference method in \cite{chen2018unified}.  
By assuming $\uvec(\x_i(t), t)\approx \dot{\x}_i(t)$, the discrete energy is 
\begin{equation}\label{eq:F_h}
\begin{aligned}
& \mathcal{F}_h\left(\{\x_i\}_{i=1}^N\right) \\
& ~~ = \frac{1}{N} \sum_{i = 1}^N \left( \ln \left( \frac{1}{N}\sum_{j = 1}^N K_h(\x_i, \x_j) \right) + V(\x_i) \right),
\end{aligned}
\end{equation}
and the discrete dissipation is
\begin{equation}\label{eq:D_h}
-2\mathcal{D}_h\left(\{\x_i\}_{i=1}^N\right)=- \frac{1}{N} \sum_{i=1}^N ||\dot{\x}_i(t))||^2,
\end{equation}
where $\dot{\x}_i(t) = \frac{\dd}{\dd t} \x_i$ is the velocity of each particle. 

We can derive the equation of $\dot{\x}_i(t)$ via a discrete energetic variational approach \cite{liu2019lagrangian}
\begin{equation}\label{dEnVar}
\frac{\delta \mathcal{D}_h}{\delta \dot{\x}_i(t)} = - \frac{\delta \mathcal{F}_h}{\delta \x_i}, 
\end{equation}
which is the energetic variational approach performed at the particle level. 
An advantage of employing the discrete energetic variational approach is that the resulting system of $\dot{\x}_i(t)$'s preserves the variational structure at the particle level. 
The benefit of this property is discussed in Remark 2 in Section 3.3. 
By direct derivation of the variations of the both sides of \eqref{dEnVar}, we obtain a systems of ODEs for $\x_i(t)$ as
\begin{equation}\label{ODE_blob}
\begin{aligned}
\dot{\x}_i(t) = & - \left( \frac{ \sum_{j=1}^N \nabla_{\x_i} K_h(\x_i, \x_j)}{\sum_{j = 1}^N K_h(\x_i, \x_j) } \right. \\
& + \left.\sum_{k=1}^N \frac{\nabla_{\x_i} K_h(\x_k, \x_i)}{\sum_{j=1}^N K_h(\x_k, \x_j)}+ \nabla_{\x_i} V(\x_i) \right), \\
& \text{for }i=1,\ldots, N.
\end{aligned}
\end{equation}
It corresponds to the ODE system of the Blob scheme proposed in \cite{chen2018unified} for ParVI. 
However, our derivation of \eqref{ODE_blob} is different from \cite{chen2018unified}. 

The {\bf Variation-then-Approximation} approach inserts the empirical approximation \eqref{emp} to \eqref{velocity}. 
Note that \eqref{velocity} is obtained after the variational step in \eqref{eq_fb}.
Thus variation step is done before the approximation step. 
Formally, the main difficulty in applying the empirical approximation \eqref{emp} to \eqref{velocity} is how to evaluate $\nabla \rho_N(\x, t)$, since $\rho_N(\x, t)$ is defined based on $\delta$ functions as in \eqref{emp}, and $\nabla \delta$ is not well defined.
One way to circumvent this difficulty is to introduce a suitable kernel regularization \cite{degond1990deterministic, lacombe1999presentation}. 
Different kernel regularization methods 
will result in different ODE systems of particles.
In the following, we show that by applying approximation to \eqref{velocity}, which is the result of variation procedure, we can obtain some existing ParVI methods. 

As pointed out in \cite{liu2017stein} and \cite{lu2019scaling}, the ODE system corresponding to the standard SVGD is
\begin{equation*}
\begin{aligned}
& \dot{\x}_i(t) = - \sum_{j=1}^N \left(K_h(\x_i, \x_j) \nabla V(\x_j) + \nabla_{\x_i} K_h(\x_i, \x_j) \right).
\end{aligned}
\end{equation*}
This ODE system can also be obtained using the EVI framework as well.
After approximating $\rho$ by $\rho_N$ in \eqref{velocity}, we can convolute to the right-hand side of \eqref{velocity} by a kernel function $K_h$ to obtain
\begin{equation*}
\rho_N(\x,t) \uvec = K_h * (\rho_N \nabla V + \nabla \rho_N(\x,t)),
\end{equation*}
which directly leads to the same ODE system as the above one of SVGD. 

Another ParVI method is the Gradient Flow with Smoothed test Function (GFSF), proposed by \cite{liu2019understanding}. 
Using the EVI framework, GFSF can be obtained by applying convolution to both sides of \eqref{velocity} with a kernel function $K_h$
\begin{equation*}
K_h * (\rho_N \uvec) = - K_h * (\rho_N\nabla V + \nabla \rho_N),
\end{equation*}
which gives us (let $K_{ij}=K_h(\x_i,\x_j)$ for short)
\begin{equation*}
\sum_{j=1}^N K_{ij} \dot{\x}_j(t) = - \sum_{j=1}^N \left( K_{ij} \nabla V(\x_j) + \nabla_{\x_i} K_h(\x_i,\x_j)\right).
\end{equation*}
Although its right-hand is exactly the descent direction in SVGD, the left is different from SVGD.

The third ParVI method we discuss is the Gradient Flow with Smoothed Density (GFSD), proposed in \cite{lacombe1999presentation, degond1990deterministic, liu2019understanding}. 
Under the EVI framework, GFSD can be obtained from $\uvec = - \nabla \rho/\rho - \nabla V$, by applying convolution to both the numerator and denominator of the first term with a kernel function $K_h$, i.e., 
\begin{equation*}
\uvec(\x) = \frac{\rho_N * \nabla K_h }{\rho_N * K_h} - \nabla V (\x). 
\end{equation*}
It leads to the same ODE system of the GFSD
\begin{equation*}
\dot{\x}_i(t) = - \left( \frac{\sum_{j=1}^N \nabla_{\x_i} K_h(\x_i, \x_j)}{\sum_{j=1}^N K_h(\x_i, \x_j)} + \nabla V(\x_i) \right). 
\end{equation*}
In SVGD and GFSF, kernel function are applied to the velocity equation directly, whereas in GFSD, the $\delta$ function in the empirical measure (\ref{emp}) is approximated by a suitable kernel $K_h(\x)$, that is
\begin{equation}\label{ker_rho}
\tilde{\rho}_N(\x,t) = \frac{1}{N} \sum_{j=1}^N K_h(\x - \x_j(t)),
\end{equation}
which is more widely used in statistics \cite{gershman2012nonparametric}.

Here we aim to show the readers that EVI is a very general framework for variational inference. 
Even if we choose a simple form of $\eta(\rho)=\rho$, exchanging the variation and approximation steps can lead to various ODE systems of the particles. 
Some of these ODE systems have already been created from other perspectives as shown above. 
But many new ParVI methods can be created, as shown in Section 3.3. 
This is an appealing advantage of the proposed EVI. 

\subsection{Explicit vs Implicit Euler}

In this subsection, we discuss how to derive a ParVI algorithm from the ODE system.
To solve the ODE system \eqref{ODE_blob} derived from the ``Approximation-then-Variation'' approach, one can use the explicit or implicit Euler method.  
Using the explicit Euler method, we obtain the following numerical scheme
\begin{equation}\label{Explict_EL}
\frac{1}{N} \frac{\x^{n+1}_i - \x^{n}_i}{\tau_n}= - \frac{\mathcal{F}_h}{\delta \x_i} \left( \{\x^{n}_i\}_{i=1}^N) \right),
\end{equation}
where $\tau_n$ is the step-size. Here $\mathcal{F}_h$ is the discrete KL-divergence defined in \eqref{eq:F_h}, and $\frac{\mathcal{F}_h}{\delta \x_i}$ is 
\begin{equation*}
\begin{aligned}
& \frac{\delta \mathcal{F}_h}{\delta \x_i} \left( \{\x^{n}_i\}_{i=1}^N) \right) =
\frac{1}{N}\left( \frac{ \sum_{j=1}^N \nabla_{\x_i} K_h(\x_i^n, \x_j^n)}{\sum_{j = 1}^N K_h(\x_i^n, \x_j^n) } \right. \\
& \quad \quad ~~ + \left.\sum_{k=1}^N \frac{\nabla_{\x_i} K_h(\x_k^n, \x_i^n)}{\sum_{j=1}^N K_h(\x_k^n, \x_j^n)}+ \nabla_{\x_i} V(\x_i^n) \right). \\
\end{aligned}
\end{equation*}
Scheme \eqref{Explict_EL} is exactly the Blob scheme proposed in \cite{chen2018unified}. 
The explicit Euler scheme is also used to solve various ODE systems 
associated with other existing ParVI methods \cite{liu2016stein, chen2018unified, liu2018riemannian, liu2019understanding}. 
To implement these methods, AdaGrad \cite{duchi2011adaptive} is often used to update the step-size. 
Although these algorithms perform well in practice, the AdaGrad scales each component of the updating direction differently. 
As a result, the updating directions of these algorithms are different from their original ODE systems. 
So the Blob scheme is equivalent to minimize the discrete energy $\mathcal{F}_h(\{\x_i \}_{i = 1}^N )$ by the AdaGrad algorithm.

An alternative approach is to adopt the implicit Euler scheme for the temporal discretization, i.e.,
\begin{equation}\label{Im_Euler}
\frac{1}{N}\frac{\x_i^{n+1} - \x_i^n}{\tau} = - \frac{\delta \mathcal{F}_h}{\delta \x_i} \left( \{ \x_i^{n+1} \}_{i=1}^N \right).
\end{equation}
The equations \eqref{Im_Euler} for $i=1,\ldots,N$ form a system of nonlinear equations.
To solve them, we first define 
\begin{equation}\label{eq:Jn}
J_n( \{ \x_i \}_{i=1}^N):= \frac{1}{2 \tau} \sum_{i=1}^N ||\x_i - \x_i^n||^2 / N + \mathcal{F}_h( \{ \x_i \}_{i=1}^N).
\end{equation}
In fact, \eqref{Im_Euler} is the gradient of $J_n(\{ \x_i \}_{i=1}^N)$ with respect to the vectorized $\{ \x_i \}_{i=1}^N$ (see the proof of Theorem \ref{thm:converge} in Appendix D).
Therefore, we can solve the nonlinear equations by solving the optimization problem.
\begin{equation}\label{Min_problem}
\{\x^{n+1}_i \}_{i=1}^N = \text{argmin}_{\{\x_i\}_{i=1}^N} J_n(\{ \x_i \}_{i=1}^N),
\end{equation} 
which is the celebrated proximal point algorithm (PPA) \cite{rockafellar1976monotone}.
The first term in \eqref{eq:Jn} can be viewed as a regularization term. 
Intuitively, when $\tau$ is relatively small, the first term can be the dominating term of $J_n( \{ \x_i \}_{i=1}^N)$ compared with $\mathcal{F}_h( \{ \x_i \}_{i=1}^N)$. 
Since it is also quadratic in $\{\x_i\}_{i=1}^N$, it can make the optimization relatively easier to solve than directly minimizing $\mathcal{F}_h( \{ \x_i \}_{i=1}^N)$. 
Besides, with a properly chosen $\tau$ value, the minimizer of \eqref{Min_problem} can lead to a small value of $\mathcal{F}_h( \{ \x_i \}_{i=1}^N)$, which is also close to $\{\x_i^n \}_{i=1}^N$.
The optimization problem \eqref{Min_problem} can be solved by a suitable nonlinear optimization. 
We can show the following convergence result. 
Its proof is in Appendix D. 
\begin{theorem}\label{thm:converge}
For a sufficiently smooth target distribution $\rho^{*}$ and any given $\{\x_i^n \}_{i=1}^N$, there exists at least one minimal solution of \eqref{Min_problem} $\{\x_i^{n+1}\}_{i=1}^N$ that also satisfies 
\begin{equation}\label{eq:decrease}
\frac{ \mathcal{F}_h( \{ \x^{n+1}_i \}_{i=1}^N) - \mathcal{F}_h(\{ \x^n_i \}_{i=1}^N)}{\tau} \leq - \frac{1}{N}\sum_{i=1}^N \frac{||\x^{n+1}_i - \x^{n}_i||^2}{2 \tau^2}.
\end{equation}
Moreover, if the series $\{\x^n_i \}_{i=1}^N$ satisfies \eqref{eq:decrease}, then $\{\x^n_i \}_{i=1}^N$ converges to a stationary point of $\mathcal{F}_h(\{ \x_i \}_{i=1}^N)$ as $n \rightarrow \infty$.
\end{theorem}

Theorem \ref{thm:converge} guarantees the existence of a solution of \eqref{Im_Euler} that also decreases the discrete KL-divergence in each iteration. 
We summarize the algorithm of using the implicit Euler scheme to solve the ODE system \eqref{ODE_blob} into Algorithm \ref{alg:EVI-Im}. 
Here MaxIter is the maximum number of iteration of the outer loop. 
\begin{algorithm}[ht]
\caption{EVI with Implicit Euler Scheme (EVI-Im)}
\label{alg:EVI-Im}
\begin{algorithmic}
 \STATE {\bfseries Input:} The target distribution $\rho^{*}(\x)$ and a set of initial particles $\{\x_i^0\}_{i=1}^N$ drawn from a prior $\rho_0(\x)$.
 \STATE {\bfseries Output:} A set of particles $\{\x_i^{*}\}_{i=1}^N$ approximating $\rho^*$.
 \FOR{$n=0$ {\bfseries to} $\textrm{MaxIter}$}
 \STATE Solve $\{\x^{n+1}_i \}_{i=1}^N = \text{argmin}_{\{\x_i\}_{i=1}^N} J_n(\{ \x_i \}_{i=1}^N)$. 
 \STATE Update $\{ \x_i^{n}\}_{i=1}^N$ by $\{ \x_i^{n+1} \}_{i=1}^N$. 
 \ENDFOR
\end{algorithmic}
\end{algorithm}

Using Algorithm \ref{alg:EVI-Im}, we update the position of particles by closely following the continuous energy-dissipation law, which provides an efficient way to push the particles to approximate the target distribution. 
In practice, it is not necessary to obtain the exact minimizer of $J_n(\{ \x_i \}_{i=1}^N)$ in each iteration.
In fact, we only need to find $\{ \x_i^{n+1}\}_{i=1}^N$ such that
\begin{equation*}
\mathcal{F}_h ( \{ \x^{n+1}_i \}) \leq \mathcal{F}_h ( \{ \x^{n}_i \}),
\end{equation*}
which usually can be achieved in a few steps via the gradient descent method or Newton-like methods with suitable step sizes to $J_n ( \{ \x_i \}_{i=1}^N )$. 
One can even adopt a line search procedure to guarantee that $J_n( \{ \x_i^{n+1} \}_{i=1}^N) \leq J_n( \{ \x^n_i \}_{i=1}^N)$.
This optimization perspective can also lead to other ParVI methods. 
For example, the idea of Stein Variational Newton (SVN) type algorithms \cite{detommaso2018stein, chen2019projected} is the same as doing one Newton step to decrease $J_n(\{\x_i\}_{i=1}^N)$.

To implement Algorithm \ref{alg:EVI-Im}, we adopt the gradient descent with Barzilai-Borwein step size \cite{barzilai1988two} to solve the optimization problem \eqref{Min_problem}.
Numerical experiments show that such algorithm usually can find a stationary point of $J_n(\{ \x \}_{i=1}^N)$ that also satisfies \eqref{eq:decrease} with relatively small value of $\tau$. 
Since it is not necessary to find the exact optimal solution of $J_n(\{ \x \}_{i=1}^N)$, especially in the early stage of the outer loop in Algorithm \ref{alg:EVI-Im}, we can fix the maximum number of iterations for the inner loop (the loop of minimizing $J_n(\{ \x \}_{i=1}^N)$) to reduce computation.

\begin{remark}
The key point in the proposed numerical algorithm is to reformulate the implicit Euler scheme into an the optimization problem \eqref{Min_problem}, which is equivalent to apply the proximal point algorithm (PPA) \cite{rockafellar1976monotone} to the discrete energy $\mathcal{F}_h(\{\x_i \}_{i=1}^N)$. 
We can decrease $\mathcal{F}_h(\{ \x_i \}_{i=1}^N)$ in each iteration and have the convergence of the algorithm at the discrete level. In other words, in each iteration, the particles are moved as they are intended by the specified dissipation law or the mechanism of decreasing the KL-divergence. 
This is the benefit of the Variation-then-Approximation approach. 
For other ParVI methods, it is unclear whether the right-hand sides of the ODEs are the gradients of some functions. 
Therefore, even though the implicit Euler scheme can be applied to these ParVI methods, the resulted ODE system can not be reformulated as an optimization problem, as we have shown in \eqref{Min_problem}.
\end{remark}

High-order temporal discretization can also be used to solve \eqref{ODE_blob}, such as the Crank-Nicolson scheme and BDF2 \cite{iserles2009first}. 
Within the variational structure at the particle level, these schemes can also be formulated into optimization problems \cite{matthes2019variational, du2019phase}.

\subsection{Choice of Kernel}

We briefly discuss the choice of kernel, or more precisely, the choice of bandwidth $h$. 
The role of kernel function $K_h(\x - \x_i)$ is essentially to approximate $\delta(\x - \x_i)$. 
Considering this role, $h$ should be as small as possible when the number of particles is large. 
However, in practice, since the number of particles is finite, it is not clear how small $h$ should be. 
Intuitively, for Gaussian kernel, $h$ controls the inter-particle distances.
In the original SVGD \cite{liu2016stein}, the bandwidth is set to be $h = \text{med}^2 / \log N$ where $\text{med}$ is the median of the pairwise distance between the current particles. The median trick updates the bandwidth after each iteration. 
However, as shown in \cite{liu2019understanding}, the median trick only works well for the SVGD. In \cite{liu2019understanding}, the authors proposed a Heat Equation based (HE) method. 
Their idea is to compute the optimal bandwidth after each iteration such that the evolution of approximated density matches the rule of the Heat Equation. 
Although the HE method works well during the numerical experiments, it requires solving an optimization problem to obtain the optimal $h$ after each iteration, which is time-consuming.
Recently, a matrix-valued kernel for SVGD has been proposed in \cite{wang2019stein}, in which some anisotropic kernels are used. The selection of kernels is based on Fisher information, i.e., the Hessian of the $V(\x)$. 
Although matrix-valued kernel works well in practice, as we have shown in Section 4.2, the computational costs will be large.  

The optimal bandwidth and the choice of kernel function are problem-dependent. 
Sometimes, a non-Gaussian kernel might be better \cite{francois2005locality}. 
We do not intend to further the discussion here.
In the examples of Section 4, we fix the bandwidth of the Gaussian kernel by conducting multiple trials. 
The results show that fixed kernel bandwidth works well in many situations for the proposed Algorithm \ref{alg:EVI-Im}.

\section{Experiments}

We present several examples that demonstrate the proposed EVI scheme summarized in Algorithm \ref{alg:EVI-Im} (or EVI-Im for short). 
The results are compared with some other deterministic ParVI methods, including AdaGrad based classical SVGD \cite{liu2016stein}, matrix-valued SVGD \cite{liu2019understanding}, and Blob method \cite{chen2018unified}. 
Additionally, we also compare our method with a gradient-based MCMC sampling method, Langevin Monte Carlo (LMC) \cite{rossky1978brownian, parisi1981correlation, roberts1996exponential} or its stochastic gradient variant, SGLD \cite{welling2011bayesian}, given by
\begin{equation*}
\x^{n+1} = \x^{n} - \epsilon_n \nabla (\log \rho^*) + \sqrt{2 \epsilon_n} \xi,
\end{equation*}
where $\xi$ is the random term $\xi \sim \mathcal{N}(0, 1)$, and $\rho^*$ is the target/posterior distribution. 

In EVI-Im, the number of iterations is ``$n$'' defined in the outer loop in Algorithm \ref{alg:EVI-Im}. 
Therefore, one iteration leads to one update of the positions of \emph{all} the particles. 
We need to point out that the amount of computation in one iteration of the outer loop of the EVI-Im method is much larger than the other ParVI methods discussed here since the optimization problem \eqref{Min_problem} needs to be solved in each iteration of EVI-Im. 
To compare the computational costs, we also show the actual CPU time of each method in Section 4.2 and 4.4.

\subsection{Toy examples via EVI-Im}
We first test the EVI-Im on three toy examples, which are widely used as benchmark tests in existing VI literature \cite{rezende2015variational, chen2018unified, liu2019understanding}. 
In all three examples, the target distributions are known up to a constant, and we can view the EVI-Im as a deterministic sampling method for these unnormalized probability densities.

\begin{figure}[ht]
\centering
\begin{overpic}[width=\linewidth]{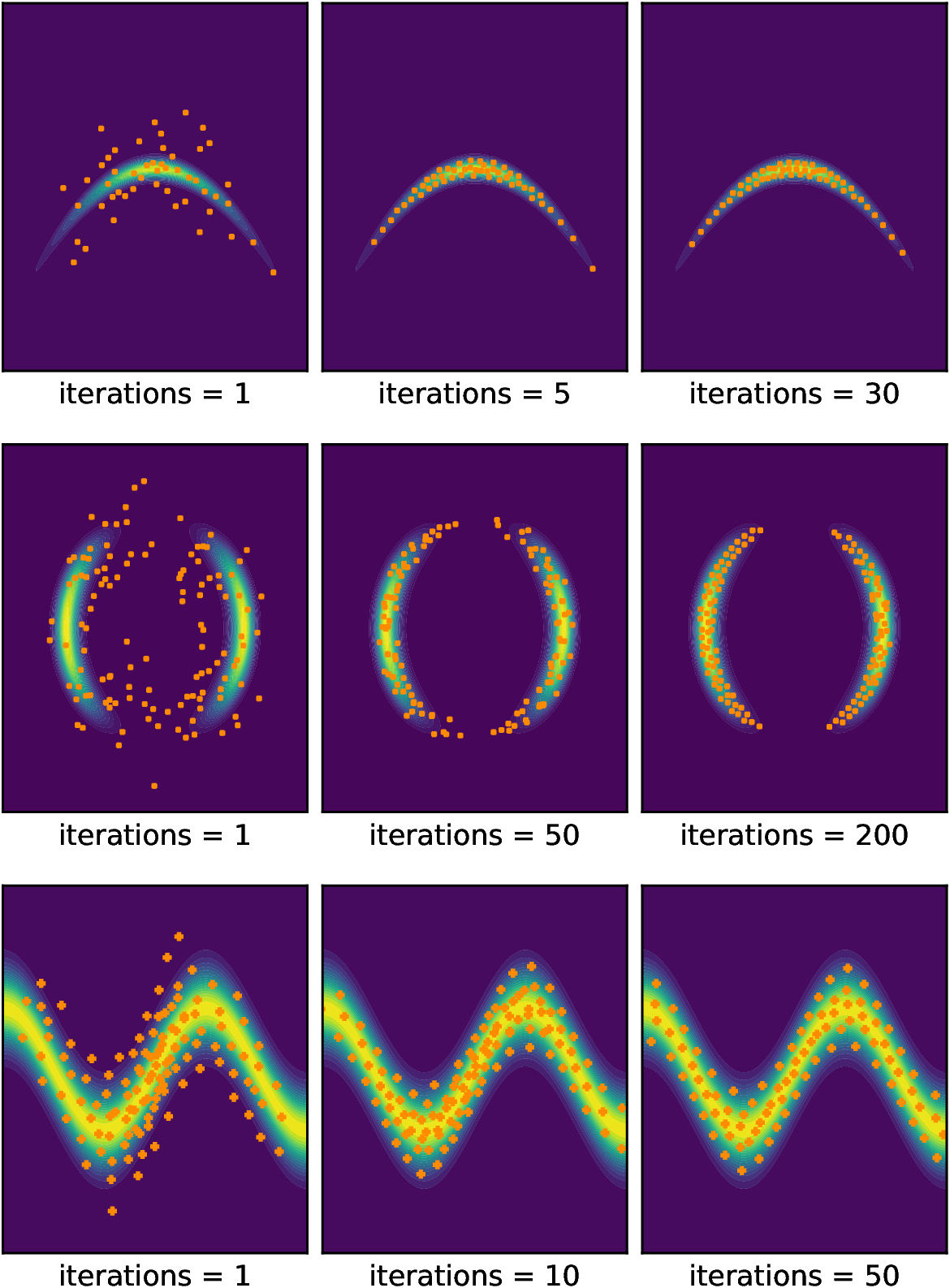}
\end{overpic}
\caption{The particles obtained by EVI-Im algorithm approximating three target distributions plotted as contours. \label{fig:toys}}
\end{figure}

The first example is modified from \cite{haario1999adaptive}. 
The target distribution is given by
\[
\rho(\x) \propto \exp\left\{ -\frac{x_1^2}{2}-\frac{1}{2}(10x_2+3x_1^2-3)^2\right\}.
\]
The second example is similar to the examples tested in \cite{rezende2015variational,liu2019understanding}, and the target distribution is
\begin{align*}
\rho(\x) \propto & \exp\left\{-2((x_1^2+x^2_2)-3)^2\right.\\
& + \left. \log\left(e^{-2(x_1-2)^2}+e^{-2(x_2+2)^2}\right) \right\},
\end{align*}
which has two components.
The third example is adapted from \cite{rezende2015variational} and \cite{chen2018unified} with the target distribution given by
\[
\rho(\x)\propto \exp\left\{-\frac{1}{2}\left[\frac{x_2-\sin(\frac{\pi x_1}{2})}{0.4}\right]^2\right\}.
\]

In all three examples, the initial particles are sampled from the two-dimensional standard Gaussian distribution. 
We use $N=50$ particles for the first example and $N=120$ particles for the second and third examples. 
The bandwidth of the kernel is $h=0.05$ for the first and second examples and $h=0.2$ for the third example. 
We set $\tau=0.01$ for all examples. 
The final results in Fig. \ref{fig:toys} show that the particles returned by the EVI-Im approximate the target distributions reasonably well. 
The second example is the most challenging one and requires more iterations because the support region (where the density is significantly larger than 0) of the target distribution is not connected and contains two banana-shaped areas. 

\subsection{Comparison on a star-shaped distribution}

\begin{figure*}[htbp]
\centering
\includegraphics[width=\linewidth]{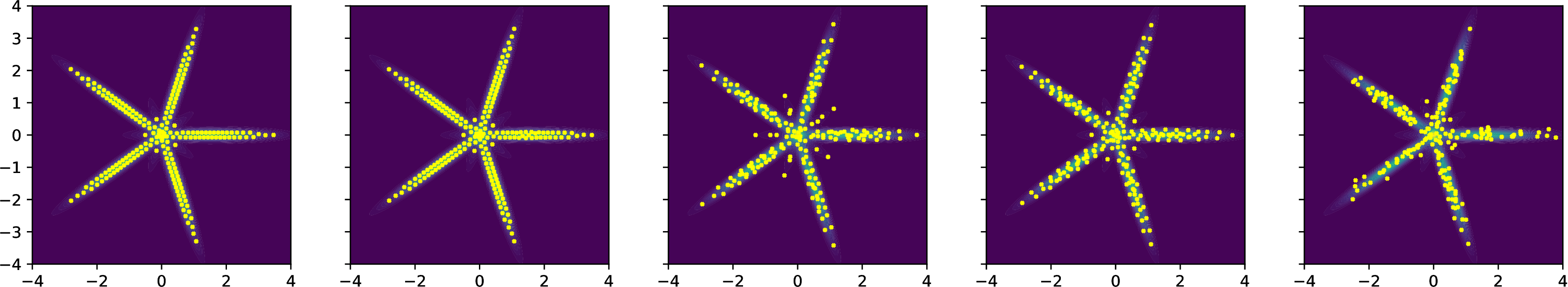}
\caption{ Particles obtained by various methods [200 particles]: (a) EVI-Im after 20 iterations, (b) Blob method (after 1000 iterations), (c) SVGD (after 1000 iterations), (d) matrix-valued SVGD (after 200 iterations) and (e) LMC (after 3000 iterations)}\label{fig:star}
\end{figure*}

The two-dimensional synthesized example studied in \cite{wang2019stein} is a challenging one, as the posterior has a star-shaped contour plot shown in Fig. \ref{fig:star}. 
We compare the EVI-Im (set $\tau = 0.5$) with the Blob method ($\mathrm{lr} = 0.5$), the classical SVGD ($\mathrm{lr} = 0.5$), the matrix-valued SVGD (mixture preconditioning matrix kernel \cite{wang2019stein}, $\mathrm{lr} = 0.5$), and the LMC method ($\epsilon_n = a(b + n)^{-c}$ with $a = 0.1$, $b=1$ and $c = 0.55$). 
The maximum number of iterations of the inner loop in EVI-Im is set to be $100$.
Here $\mathrm{lr}$ stands for the learning rate. 
In all five methods, we use $N=200$ particles and the same initial set of particles sampled from the two-dimensional standard Gaussian distribution. 
For the EVI-Im and the Blob method, we fix the kernel bandwidth to be $h=0.1$. 
The bandwidth matrix in the matrix-valued SVGD is set as the exact Hessian matrices as in \cite{wang2019stein}. 
To compare the fidelity of the particles to the target distribution, we compute the squared Maximum Mean Discrepancy (MMD$^2$) defined as \cite{arbel2019maximum}
\begin{equation*}
\begin{aligned}
{\rm MMD}^2 & = \frac{1}{N^2} \displaystyle \sum_{i, j = 1}^{N} k(\x_i, \x_j) + \frac{1}{M^2} \displaystyle \sum_{i, j = 1}^{M} k(\y_i, \y_j) \\
& - \frac{2}{N M} \displaystyle \sum_{i=1}^N \sum_{j=1}^{M} k(\x_i, \y_j) \\
 \end{aligned}
\end{equation*}
with a polynomial kernel $k(\x, \y) = (\x^\top \y/3 + 1)^{3}$, where $\{\x_i\}_{i=1}^N$ are $N=200$ particles generated by the different methods, and $\{\y_j\}_{j=1}^M$ are 5000 samples that generated from $\rho^*$ directly. 

\begin{figure}[htbp]
\centering
\includegraphics[width=0.48\linewidth]{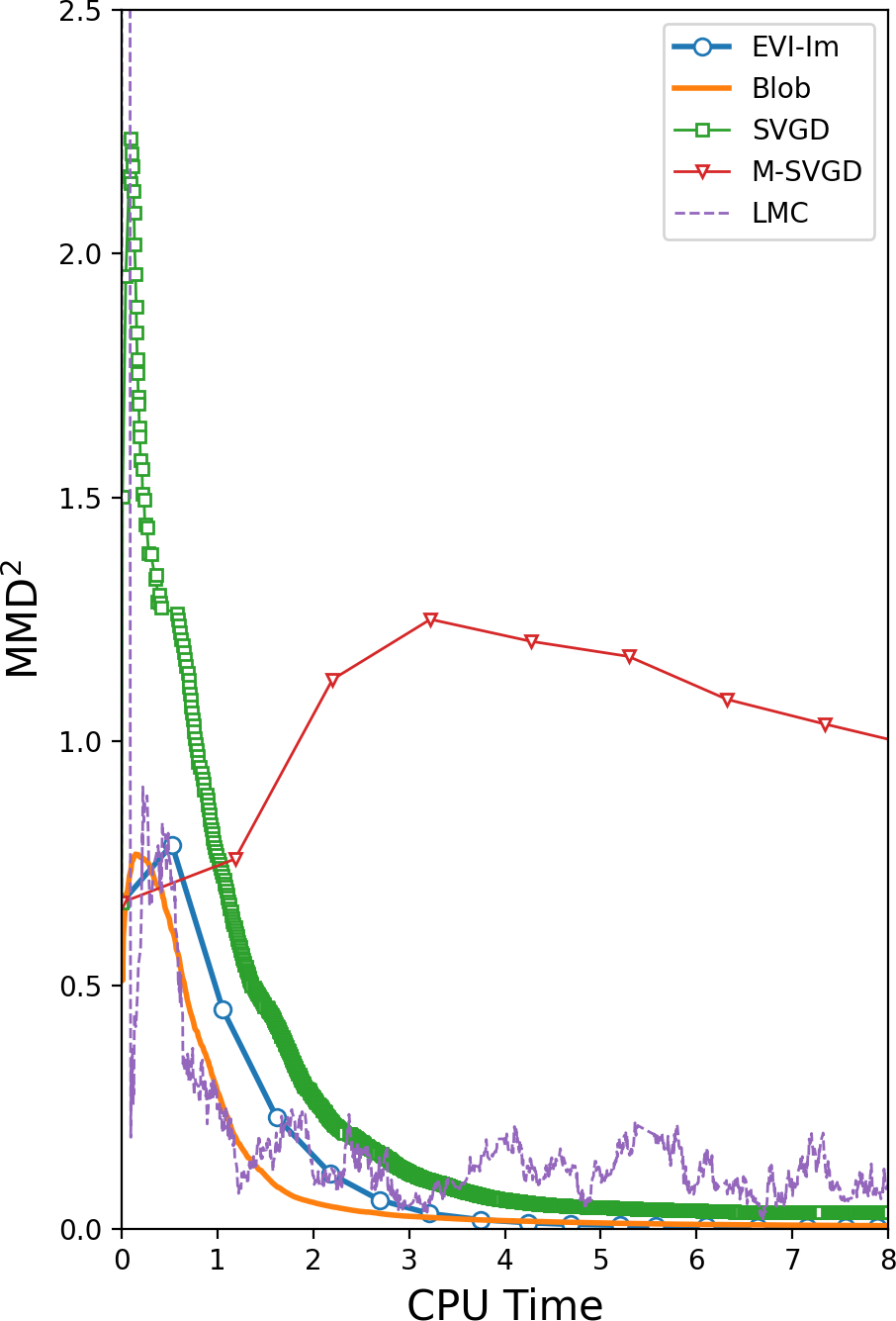}
\hfill
\includegraphics[width= 0.48 \linewidth]{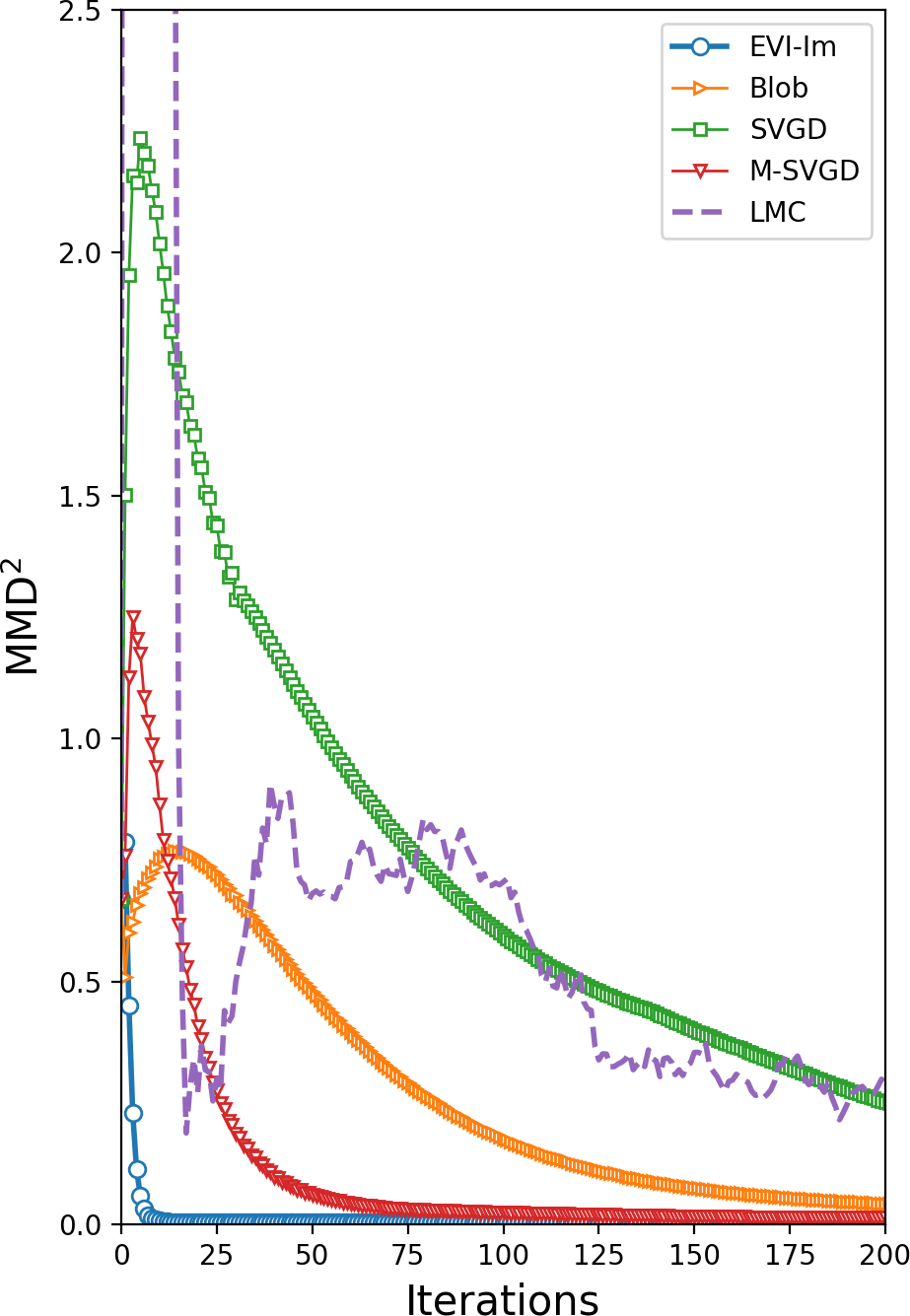}  
\caption{MMD$^2$ of each method with respect to CPU time and number of iterations.}
\label{fig:MMD}
\end{figure}

The sampling results returned by different methods are shown in Fig. \ref{fig:star}. 
The MMD$^2$ of each method with respect to the CPU time and the number of iterations is shown in Fig. \ref{fig:MMD}. 
We observe that the results returned by the EVI-Im and the Blob method are most similar compared to the other methods. 
As we mentioned earlier, they minimize the same discrete KL-divergence defined in \eqref{eq:F_h} with different optimization methods. 
The EVI-Im uses PPA whereas the Blob uses AdaGrad. 
The CPU time for both approaches is also comparable. 
The particles returned by the EVI-Im, the Blob, and the Matrix-valued SVGD appear to align more regularly than those returned by the standard SVGD. 
Moreover, the particles returned by the three methods are more likely to be concentrated in high probability areas, compared with SVGD and LMC.
Using the EVI-Im, we can obtain a good approximation within less than 20 iterations with $\tau = 0.5$. 
However, since it solves an optimization problem in each iteration, the total CPU time is slightly larger than the Blob method. 
All methods have similar computational efficiency to the LMC in terms of MMD$^2$ vs CPU time, except for the matrix-valued SVGD. 
Its computational cost increases dramatically for computing anisotropic kernels in each iteration.
We should emphasize that the total CPU time is sensitive to the choice of learning rate. 
The learning rate presented here is chosen to have the best performances according to our tests.
For EVI-Im, a slightly large time step-size $\tau$ value is preferred for computational efficiency, but the robustness of the algorithm requires a relatively small $\tau$.

\subsection{Mixture Model}

\begin{figure}[htbp]
\centering
\includegraphics[width=\linewidth]{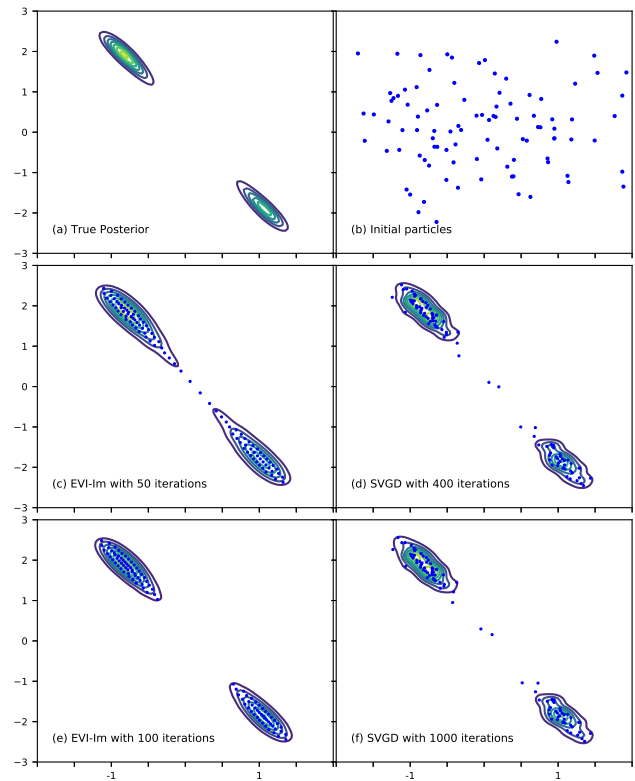}
\caption{Comparison of EVI-Im and the classic SVGD ($\text{lr}=1)$ at different stages of iterations. \label{fig:GMM}}
\end{figure}

In this subsection, we consider an example of a simple but interesting mixture model, which is studied in \cite{dai2016provable} and \cite{welling2011bayesian}.
We sample 1000 observed data from ${\bm y}_i \sim \frac{1}{2}(N(\omega_1,\sigma^2)+N(\omega_1+\omega_2,\sigma^2))$, where $(\omega_1, \omega_2) = (1,-2)$ and $\sigma = 2.5$. 
Using the prior $\omega_1, \omega_2 \sim N(0,1)$, the posterior distribution is known except the constant, which is the marginal distribution of the data. 
But it is easy to obtain its two modes, $(1,-2)$ and $(-1 ,2)$. 
The contour plot of the posterior distribution up to the constant is in Fig. \ref{fig:GMM} (a).

\begin{figure*}[htbp]
 \centering
 \includegraphics[width=\linewidth]{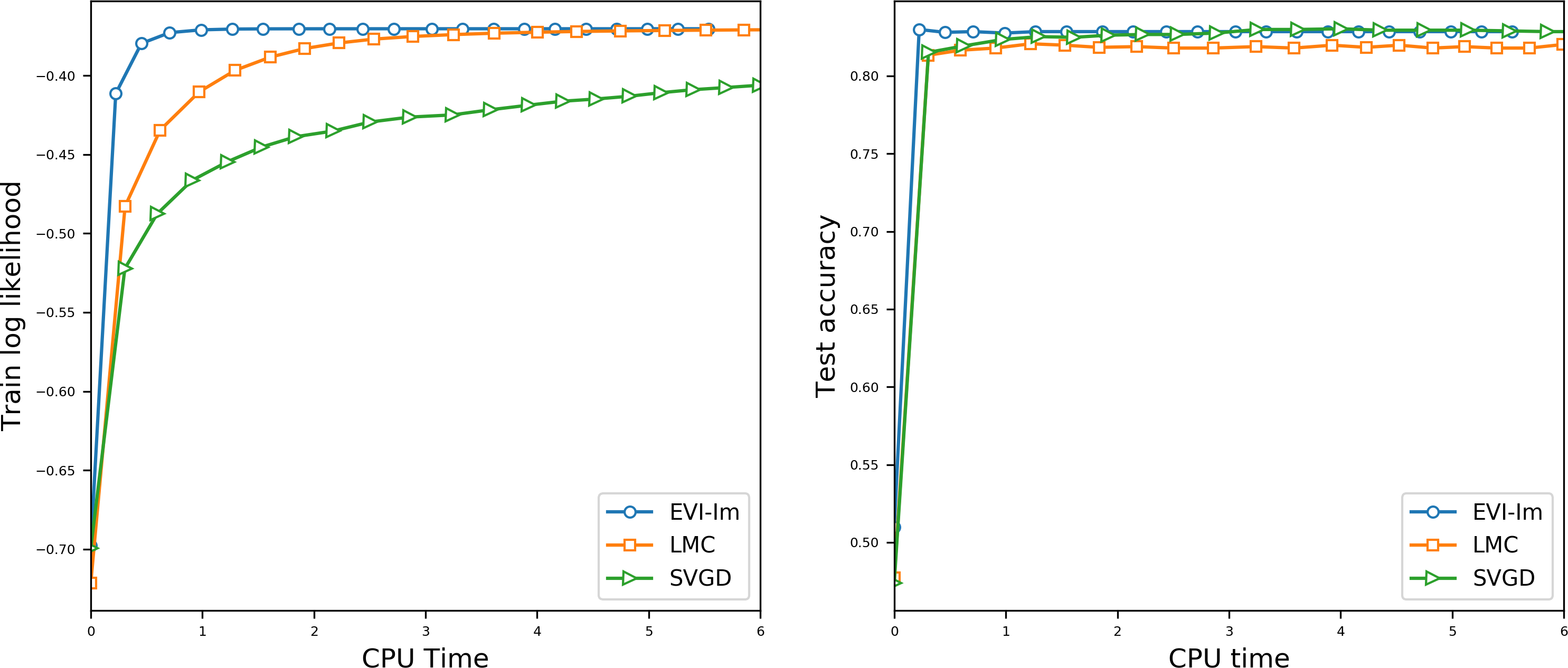}
 \caption{The log-likelihood of the training data and test accuracy of the SPLICE dataset returned by EVI-Im, SVGD and LMC methods.}
 \label{fig:lr}
 \end{figure*}

Fig. \ref{fig:GMM} shows the posterior distribution approximated by EVI-Im and SVGD ($\textnormal{lr} = 1.0$). We have tried the SVGD with learning rate $\textnormal{lr} = 0.01, 0.1, 0.5, 1.0$ and choose the best learning rate $\textnormal{lr}=1$.
For the EVI-Im, we set $\tau=0.01$. 
The same $N=100$ initial particles sampled from the prior are used in both methods, as shown in Fig. \ref{fig:GMM} (b).
Kernel density estimation with optimal bandwidth selected via cross-validation is used to generate the estimated posterior distribution for both methods. 
It also shows the approximated distributions of EVI-Im and SVGD at different iterations. 
When both methods converge, EVI-Im (100 iterations) approximates the true posterior distribution better than the SVGD (1000 iterations). 
During the iterations, the particles returned by the EVI-Im also appear to be aligned more regularly. 
But among the particles returned by SVGD, some are clustered but some are scattered widely. 
In \cite{dai2016provable}, the authors compared many other methods, such as Gibbs sampling, SGLD, and the one-pass sequential Monte Carlo (SMC). 
Compared to numerical results in \cite{dai2016provable}, the EVI-Im has better results than Gibbs sampling, SGLD, and SMC, and is also comparable to the particle mirror descent algorithm proposed in \cite{dai2016provable}.

\subsection{Bayesian Logistic Regression with Real Data Sets}

In this subsection, we apply EVI-Im to Bayesian logistic regression models. 
We first consider a small data set SPLICE (1,000 training entries, 60 features), a benchmark data set used in \cite{mika1999fisher}.  
Given the data set $\{\bm c_t,y_t\}_{t=1}^{1000}$, the logistic regression model is $p(y_t=1|\bm c_t,\bm{\omega})=[1+\textnormal{exp}(-\bm{\omega}^T \bm c_t)]^{-1}$.
The unknown parameters $\bm \omega$ are the regression coefficients, whose prior is $N(\bm{\omega};{\bf 0}, \alpha {\bf I})$.
We compare the EVI-Im method with the classic SVGD and the LMC. 
We use $N=20$ particles for each method. 
The learning rate in SVGD is set to be $0.1$. 
For LMC, we take $\epsilon_n = a(b + n)^{-c}$ with $a = 10^{-4}$, $b = 1$ and $c = 0.55$. 
For EVI-Im, we take $\tau = 0.01$ and set the maximum number of iteration in the inner loop to be $50$. 
We should emphasize these parameters may not be optimal for all the methods.
Fig. \ref{fig:lr} shows the log-likelihood of the training data and test accuracy for all methods with respect to the CPU time.
Although the test accuracies of the three methods are similar and the EVI-Im has a slight advantage, the EVI-Im is shown to achieve a larger log-likelihood with less CPU time of the training data.

We also apply the EVI-Im to a large data set Covertype \cite{wang2019stein}, which contains $581,012$ data entries and 54 features, and compare the proposed EVI-Im algorithm and the original SVGD method.
The prior of the unknown regression coefficients is chosen to be $p(\bm{\omega}) = N(\bm{\omega};0, {\bf I})$. 
Due to the large size of the data, the computation of log-likelihood $\nabla \ln \rho^*$ is expensive. 
Hence, we randomly sample a batch of data to compute a stochastic approximation of $\nabla \ln \rho^*$. 
The batch size is set to be 256 for all methods. 
Recall that in the EVI-Im algorithm, we need to solve a minimization problem to update the positions of the particles in each iteration of the outer loop. 
Since we only estimate $\nabla \ln \rho^*$ using a subset of the complete data, which is only an approximation of the exact estimate using the complete data, the EVI-Im algorithm does not need to achieve the exact local optimality in each iteration.
Thus, we choose the stochastic gradient descent method AdaGrad \cite{duchi2011adaptive} with learning rate $\mathrm{lr} = 0.1$ to minimize $J_n(\{ \x_i \}_{i=1}^N)$. 
We set the maximum number of iterations for the inner loop of AdaGrad to be 100 in the EVI-Im algorithm. 
Meanwhile, the time step-size, $\tau$, is set to be $0.1$ in the EVI-Im algorithm. 
For the SVGD method, we choose the 
best learning rate among $\mathrm{lr}=$0.01, 0.05, 0.1, 0.5, 1.0.
For all methods, we use $N=20$ particles, as in \cite{wang2019stein}.

\begin{figure}[htbp]
 \centering
 \begin{overpic}[width= 0.95 \linewidth]{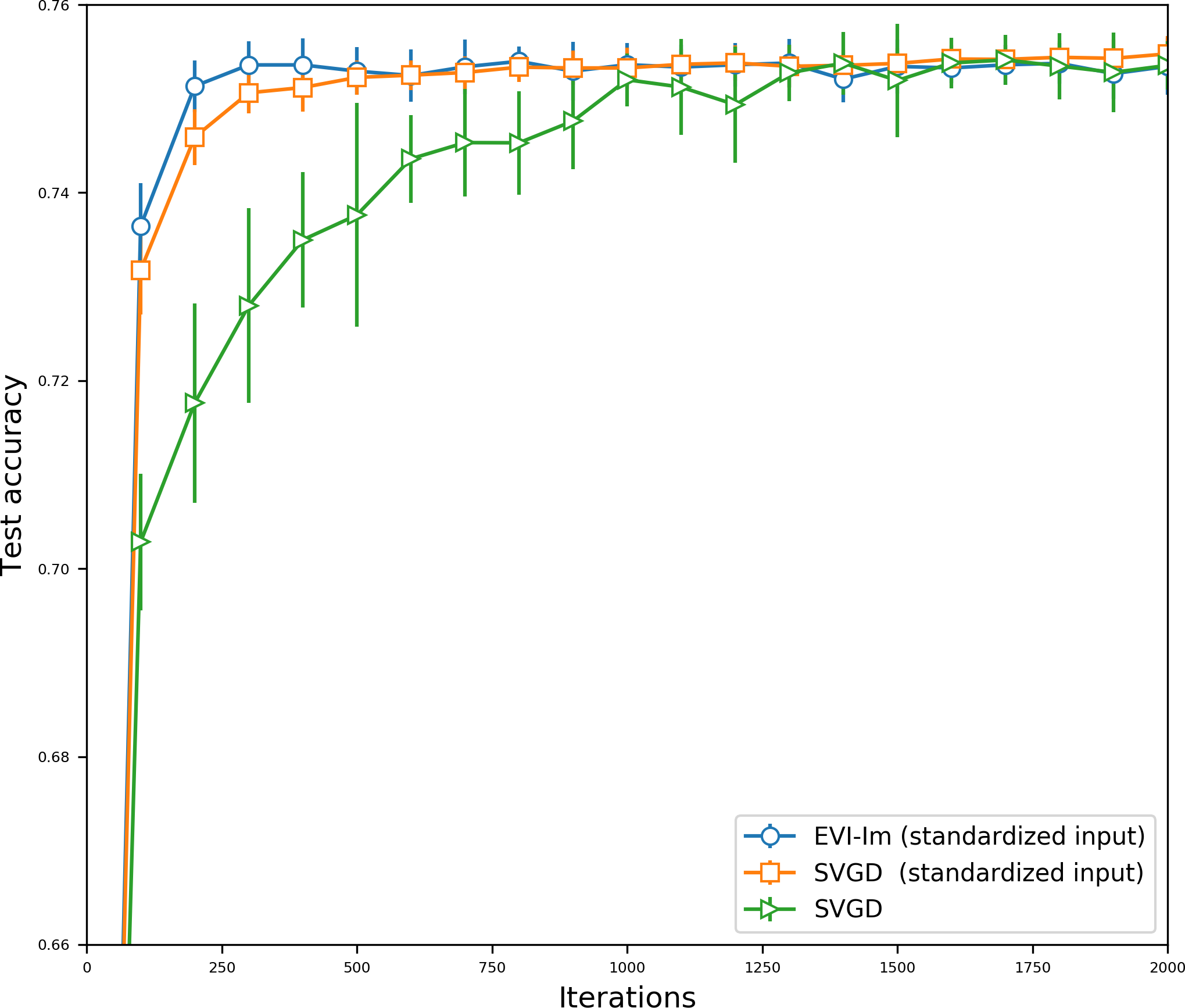}
 \put(-3, 82){(a)}
 \end{overpic}

 \vspace{1em}
 \begin{overpic}[width= 0.95 \linewidth]{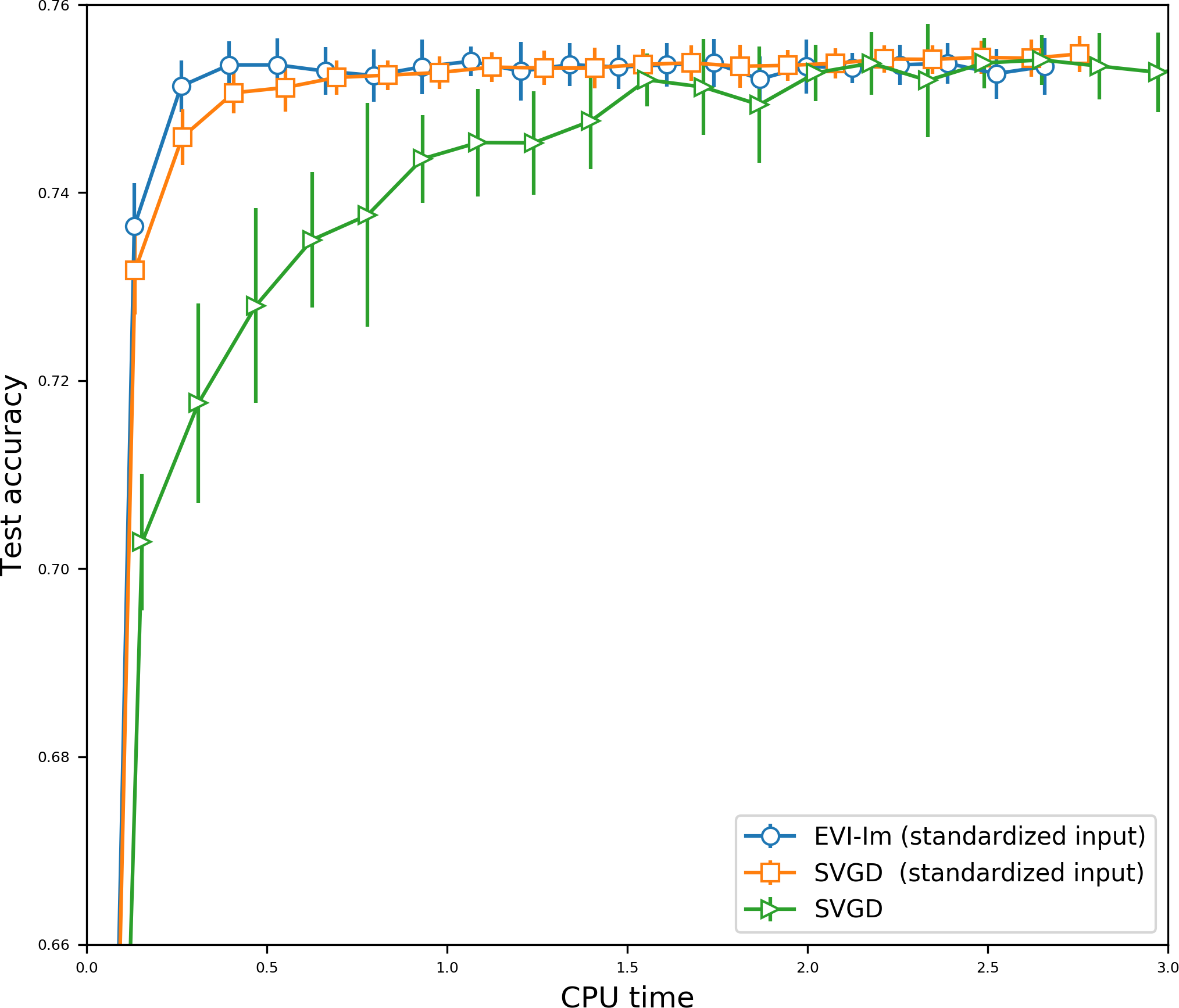}
 \put(-3, 82){(b)}
\end{overpic}
 \caption{The test accuracy of 20 simulations for Bayesian logistic regression on Covertype dataset using different methods with respect to (a) the number of iterations and (b) CPU time (in seconds). The number of iterations for EVI-Im is defined as No. of Outer Iterations $\times$ 100(No. of Inner Iterations). The error bar in each curve corresponds to the standard deviation of 20 simulations. \label{fig:lr_cov}}
 \end{figure}

In the statistical analysis of real data, it is a common practice to standardize all columns of inputs via their individual mean and standard deviation in the preprocessing stage. 
Thus, we apply both the EVI-Im and the SVGD algorithms (with $\mathrm{lr} = 0.1$) to the standardized data. 
We also apply the SVGD (with $\text{lr} = 1$) to the non-standardized data, which was done in the same way as in \cite{wang2019stein}. 
The SVGD is implemented using the codes\footnote{available from https://github.com/dilinwang820/Stein-Variational-Gradient-Descent.} by \cite{wang2019stein}. 
For each method, we have run a total of 20 simulations.
In each simulation, we randomly partition the data into training (80\% of the whole) and testing (20\% of the whole) sets. 
Fig. \ref{fig:lr_cov} shows the test accuracy of the classification of EVI-Im and SVGD applied to standardized data and SVGD applied to non-standardized data. 
The test accuracy is the average of 20 simulations. 
The CPU time of each point in (b) of Fig. \ref{fig:lr_cov} is the average CPU time of 20 simulations of every 100 AdaGrad steps for all three methods under comparison.

For the SVGD (both versions), the number of iterations counts the iterations of the single layer of loop. 
For the EVI-Im, there are two layers of loops. 
The outer loop is the for-loop in Algorithm \ref{alg:EVI-Im} and the inner loop is for the AdaGrad algorithm. 
As mentioned above, the inner loop of AdaGrad has 100 iterations. 
To compare it with SVGD, the number of iterations for EVI-Im in Fig. \ref{fig:lr_cov} (a) is defined as 
\[\text{No. of Outer Iterations }\times 100(\text{No. of Inner Iterations}).\] 
Since both methods use the AdaGrad with the same batch size, both methods conduct a very similar amount of computation in each iteration, which is confirmed by the close resemblance between (a) and (b) of Fig. \ref{fig:lr_cov}.
The proposed EVI-Im is the best among the three. 
We can also compare the EVI-Im algorithm with the matrix-valued SVGD. 
As shown in \cite{wang2019stein}, the matrix-valued SVGD can reach an accuracy of 0.75 in less than 500 iterations. 
Using the EVI-Im algorithm with standardized data, we can reach the same accuracy of 0.75 around 200 iterations.

From Fig. \ref{fig:lr_cov}, we can first conclude that standardization significantly improves the accuracy and reduce the variance of the SVGD method. 
This is expected because standardization is essentially applying different bandwidth values to different input dimensions inside the kernel function. 
As a result, the original SVGD with standardization performs similarly to the matrix-valued SVGD proposed in \cite{wang2019stein}, although the latter also linearly transforms the SVGD direction by multiplying a preconditioning matrix on the original SVGD direction.
For the same reason, EVI-Im algorithm also benefits from standardization, as it is also a kernel-based method.

At last, we point out that the proposed EVI-Im algorithm, the SVGD with or without standardized data, and the matrix-valued SVGD method \cite{wang2019stein} have similar performance when they reach convergence. 
A major reason is that due to the large size of the data, the KL-divergence is entirely dominated by the log-likelihood. 
Consequently, the interactions between particles play little effect in the updating of the particles. 
Thus, there is no significant distinction between different methods when they all reach convergence.

\section{Conclusion}
In this paper, we introduce a new variational inference framework, called \emph{energetic variational inference} (EVI), in which the procedure of minimizing VI object function is characterized by an \emph{energy-dissipation law}. A VI algorithm can be obtained by employing an energetic variational approach and proper discretization.
The EVI is a general framework. 
By specifying different components of EVI, we can derive many ParVI algorithms. 
These components include 
\begin{itemize}
\item the continuous energy-dissipation law, such as \eqref{KL_D} and \eqref{eq:anotherlaw}; 
\item the order of approximation and variation steps; 
\item numerical schemes or optimization techniques, such as implicit and explicit Euler, first-order and higher-order temporal discretization, etc. 
\end{itemize}
We have shown that some combinations of these choices lead to some existing ParVI methods. 
But many new methods can be created as such. 
In particular, by using the ``Approximation-then-Variation'' order, we can derive a particle system that inherits the variational structure from the original energy-dissipation law.
Numerical examples show that the proposed method has comparable performance with the latest ParVI methods. 
Another significant aspect is that the EVI framework is not restricted to KL-divergence, and it can be used to minimize other discrepancy measures on the difference between two distributions, such as $f$-divergence \cite{ali1966general}. 
This opens doors to many varieties in the variational inference literature.
The codes and data for all examples are available from Github \url{https://github.com/SimonKafka/EVI}. 

\section*{Acknowledge}
Y. Wang and C. Liu are partially supported by the National Science Foundation grant DMS-1759536. 
L. Kang is partially supported by the National Science Foundation grant DMS-1916467. 

\begin{appendices}
\section{Energetic variational approach}
In this appendix, we gives a brief introduction to the energetic variational approach. 
We refer interested readers to \cite{liu2009introduction, giga2017variational} for a more comprehensive description. 

As mentioned previously, the energetic variational approach provides a paradigm to determine the dynamics of a dissipative system from a prescribed energy-dissipation law, which shifts the main task in the modeling of a dynamic system to the construction of the energy-dissipation law.  
In physics, an \emph{energy-dissipation law},
yielded by the first and second Law of thermodynamics \cite{giga2017variational},
is often given by
\begin{equation}\label{EDL}
\frac{\dd}{\dd t} (\mathcal{K} + \mathcal{F}) [{\bm \phi}] = - 2 \mathcal{D}[{\bm{\phi}}, \dot{\bm \phi}],
\end{equation}
where ${\bm \phi}$ is the state variable, $\mathcal{K}$ is the kinetic energy, $\mathcal{F}$ is the Helmholtz free energy, and $2 \mathcal{D}$ is the energy-dissipation.
 If $\mathcal{K} = 0$, one can view \eqref{EDL} as a generalization of gradient flow \cite{hohenberg1977theory}. 
 
The Least Action Principle states that the equation of motion for a Hamiltonian system can be derived from the
variation of the action functional $\mathcal{A} = \int_{0}^T (\mathcal{K} - \mathcal{F})\dd \x$ with respect to ${\bm \phi}(\z, t)$ (the trajectory), i.e.,
$$
\begin{aligned}
\delta \mathcal{A} & = \lim_{\epsilon \rightarrow 0} \frac{\mathcal{A}[{\bm \phi} + \epsilon \delta {\bm \psi}] - \mathcal{A}[{\bm \psi}] }{\epsilon} \\
& = \int_{0}^T \int_{\mathcal{X}^t} (f_{\rm inertial} - f_{\rm conv}) \cdot \delta {\bm \psi} \dd \x \dd t. \\
\end{aligned}
$$
This procedure yields the conservative forces of the system, that is $ (f_{\rm inertial} - f_{\rm conv}) = \frac{\delta \mathcal{A}}{\delta {\bm \phi}}$.
Meanwhile, according to the MDP, the dissipative force can be obtained by minimizing the dissipation functional with respect to the ``rate'' $\dot{\bm \phi}$, i.e.,
$$ \delta \mathcal{D} = \lim_{\epsilon \rightarrow 0} \frac{\mathcal{D}[\dot{\bm \phi} + \epsilon \delta {\bm \psi}] - \mathcal{D}[\dot{\bm \phi}] }{\epsilon} = \int_{\mathcal{X}^t} f_{\rm diss} \cdot \delta \psi \dd \x,$$
or $f_{\rm diss} = \frac{\delta \mathcal{D}}{\delta \dot{\bm \phi}}$. According to the Newton's second law ($F = ma$), we have the force balance condition $f_{\rm inertial} = f_{\rm conv} + f_{\rm diss}$ ($f_{\rm inertial}$ plays role of $ma$), which defines the dynamics of the system
\begin{equation}\label{DF1}
\frac{\delta \mathcal{D}}{\delta \dot{\bm \phi}} = \frac{\delta \mathcal{A}}{\delta {\bm \phi}}.  
\end{equation}
In the case that $\mathcal{K} = 0$, we have
\begin{equation}\label{DF}
 \frac{\delta \mathcal{D}}{\delta \dot{\bm \phi}} = - \frac{\delta \mathcal{F}}{\delta {\bm \phi}},
\end{equation}
Notice that the free energy is decreasing with respect to the time when $\mathcal{K} = 0$. 
As an analogy, if we consider VI objective functional as $\mathcal{F}$, (\ref{EDL}) gives a continuous mechanism to decrease the free energy, and \eqref{DF1} or \eqref{DF} gives the equation ${\bm \phi}(\z, t)$.

\section{Derivation of the transport equation}

 The transport equation can be derived from the conservation of probability mass directly. Let
 \begin{equation}\label{def_F}
  \F(\z, t) = \nabla_{\z} \bm{\phi}(\z, t)
 \end{equation}
 be the deformation tensor associate with the flow map $\phi(\z, t)$, i.e., the Jacobian matrix of ${\bm \phi}$, then due to the conservation of probability mass, we have
 \begin{equation*}
  \begin{aligned}
  0 & = \frac{\dd}{\dd t} \int_{\mathcal{X}^t} \rho(\x, t) \dd \x = \frac{\dd}{\dd t} \int_{\mathcal{X}^0} \rho(\phi(\z, t), t) \det (\F(\z, t)) \dd \z \\
  & = \int_{\mathcal{X}^0} \left( \dot{\rho} + \nabla \rho \cdot \uvec + \rho (\F^{-\rm T} : \frac{\dd \F}{\dd t}) \right) \det \F \dd \z \\
  & = \int_{\mathcal{X}^t} \left( \dot{\rho} + \nabla \rho \cdot \uvec + \rho (\nabla \cdot \uvec) \right) \dd \x = 0, \\
  \end{aligned}
 \end{equation*}
 which implies that
 $$\dot{\rho} + \nabla \cdot (\rho \uvec) = 0.$$ 
Here the operation ``$:$'' between two matrix, ${\sf A} : {\sf B} = \sum_{i} \sum_{j} A_{ij} B_{ij}$ is the Frobenius inner product between two matrices ${\sf A}, {\sf B} \in \mathbb{R}^{n \times m}$.

\section{Computation of Equation (2.10)}
In this part, we give a detailed derivation of the variation of $\KL(\rho_{[\bm \phi]} | \rho^{*})$ with respect to the flow map $\bm{\phi}(\z): \mathcal{X}^0 \rightarrow \mathcal{X}^t$. Consider a small perturbation of $\bm{\phi}$
\[\bm{\phi}^{\epsilon}(\z):= \bm{\phi}(\z) + \epsilon \bm{\psi}(\z), \]
where $\bm{\psi}(\z) = \widetilde{\bm{\psi}}(\bm{\phi}(\z))$ is a smooth map satisfying 
\[ \widetilde{\bm{\psi}} \cdot \bm{\nu} = 0, \quad \text{on} ~~ \pp \mathcal{X}^t \]
with $\bm{\nu}$ be the outward pointing unit normal on the boundary, $\pp \mathcal{X}^t$. 
Thus, $\widetilde{\bm{\psi}}=\bm \psi(\bm \phi^{-1}(\x))$ and the above condition indicates that $\widetilde{\bm\psi}$ is diffused to zero at the boundary of $\mathcal{X}^t$.
For $\mathcal{X}^0 = \mathcal{X}^d = \mathbb{R}^d$, $\widetilde{\bm{\psi}} \in C_0^{\infty} (\mathbb{R}^d)$. 
We denote $\F$ as the Jacobian matrix of $\bm \phi$, i.e., $F_{ij}=\frac{\partial \phi_i}{\partial z_j}$, and $\F^{\epsilon}$ is the Jacobian matrix of $\bm \phi^{\epsilon}$, i.e., 
\[\F^{\epsilon}:= \nabla_{\z} \bm{\phi} + \epsilon \nabla_{z} \bm{\psi}.\]
Then we have
\begin{equation}\label{Var_phi}
 \begin{aligned}
 & \frac{\dd}{\dd \epsilon} \Big|_{\epsilon = 0} \KL(\rho_{[\bm \phi^{\epsilon}]} || \rho^{*}) \\
 & = \frac{\dd}{\dd \epsilon} \Big|_{\epsilon =0 }\left(\int_{\mathcal{X}^0} \frac{\rho_0}{\det (\F^{\epsilon})} \ln \left(\frac{\rho_0}{\det (\F^{\epsilon})} \right) \det(\F^{\epsilon})\dd \z\right. \\
 & \left.\quad \quad \quad \quad + \int_{\mathcal{X}^0} \frac{\rho_0}{\det (\F^{\epsilon})} V( \bm{\phi}^{\epsilon}(\z)) ~\det(\F^{\epsilon}) \dd \z\right) \\
 & = \frac{\dd}{\dd \epsilon} \Big|_{\epsilon =0 }\left(\int_{\mathcal{X}^0} \rho_0\ln \rho_0-\rho_0\ln \det \F^{\epsilon} + \rho_0 V( \bm{\phi}^{\epsilon}(\z)) \dd \z\right) \\
  & =\int_{\mathcal{X}^0} -\rho_0\frac{\dd}{\dd \epsilon} \Big|_{\epsilon =0 }\left(\ln \det (\F^{\epsilon}) + V( \bm{\phi}^{\epsilon}(\z))\right) \dd \z \\
 & = \int_{\mathcal{X}^0} - \rho_0 (\F^{-\top}:\nabla_{\z} \bm{\psi}) + (\nabla_{\x} V \cdot \bm{\psi}) \rho_0  \dd \z. 
 \end{aligned}
\end{equation}
For two matrices of the same size, define ${\sf A}:{\sf B}=\sum_i\sum_j A_{ij}B_{ij}=\tr({\sf A}^\top {\sf B})$. 
Since 
\[
\frac{\dd \det(\F^{\epsilon})}{\dd \epsilon}=\det(\F^{\epsilon})\tr\left[ (\F^{\epsilon})^{-1}\frac{\dd \F^{\epsilon}}{\dd \epsilon}\right],
\]
we have 
\begin{align*}
\frac{\dd \det(\F^{\epsilon})}{\dd \epsilon}\Big|_{\epsilon =0 }&=\det(\F)\tr\left[\F^{-1}\nabla_{\z}\bm \psi\right]\\
&=\det(\F)(\F^{-\top}:\nabla_{\z}\bm \psi).
\end{align*}
Hence we have the last result in \eqref{Var_phi}. 

Based on the definition of $\bm \phi$, we have the following. 
\begin{align*}
&\x =\bm \phi(\z), \quad \z=\bm \phi^{-1}(\x)\\
&\bm \psi(\z)=\widetilde{\bm \psi}(\bm \phi(\z))=\widetilde{\bm \psi}(\x)\\
&\rho(\x)=\rho_0(\bm \phi^{-1}(\x))\det(\nabla_{\x}\bm \phi^{-1}(\x))\\
&\rho_0(\z)=\rho_0(\bm \phi^{-1}(\x))=\frac{\rho(\x)}{\det(\nabla_{\x}\bm \phi^{-1}(\x))}.
\end{align*}
The second summand of \eqref{Var_phi} becomes
\begin{align*}
&\int_{\mathcal{X}_0}\rho_0\left[(\nabla_{\x}V)^\top \bm \psi\right]\dd \z\\
=&\int_{\mathcal{X}_t}\frac{\rho(\x)}{\det(\nabla_{\x}\bm \phi^{-1}(\x))} \left[(\nabla_{\x}V)^\top \bm \psi\right]\dd \bm \phi^{-1}(\x)\\
=&\int_{\mathcal{X}_t}\frac{\rho(\x)}{\det(\nabla_{\x}\bm \phi^{-1}(\x))} \left[(\nabla_{\x}V)^\top \bm \psi\right]\det(\nabla_{\x}\bm \phi^{-1}(\x)) \dd \x\\
=&\int_{\mathcal{X}_t}\rho(\x)\left[(\nabla_{\x}V)\cdot \bm \psi\right]\dd \x.
\end{align*}

Now we investigate the first summand in \eqref{Var_phi}. 
Based on the definition of $\widetilde{\bm \psi}$, we can see that 
\begin{align*}
(\nabla_{\x} \widetilde{\bm \psi})_{i,j}&=\sum_{k=1}^d \frac{\partial \psi_i}{\partial z_k}\cdot \frac{\partial z_k}{\partial x_j}=\sum_{k=1}^d (\nabla_{\z}\bm \psi)_{i,k}(\nabla_{\x}\bm \phi^{-1})_{k,j}\\
\nabla_{\x} \widetilde{\bm \psi}&=\nabla_{\z}\bm \psi \left[\nabla_{\x}\bm \phi^{-1} \right]^\top=(\nabla_{\z}\bm \psi) \F^{-\top},
\end{align*}
because $\F=\nabla_{\z}\bm \phi(z)=\left(\frac{\partial x_i}{\partial z_j}\right)_{i,j}$, $\F^{-1}=\left(\frac{\partial z_i}{\partial x_j}\right)_{i,j}=(\nabla_{\x} \bm \phi^{-1}(\x))^{-1}$. 
Divergence of $\widetilde{\bm \psi}(\bm x)$ is 
\begin{align*}
&\nabla_{\x} \cdot \widetilde{\bm \psi}(\bm x)=\sum_{i=1}^d \frac{\partial \tilde{\psi}_i}{\partial x_i}=\tr(\text{Jacobian of }\widetilde{\bm \psi})\\
=&\tr(\nabla_{\x} \widetilde{\bm \psi})=\tr((\nabla_{\z}\bm \psi) \F^{-\top})=\tr(\F^{-1}\nabla_{\z}\bm \psi).
\end{align*}
Therefore, the first summand in \eqref{Var_phi} becomes, 
\[-\int_{\mathcal{X}_0}\rho_0 (\F^{-\top} : \nabla_{\z} \bm{\psi})\dd z
=-\int_{\mathcal{X}_t}\rho(\x) (\nabla_{\x}\cdot \widetilde{\bm \psi})\dd \x. 
\]
Following the corollary of Divergence theorem, 
\begin{align*}
& \int_{\mathcal{X}_t}\rho(\x) (\nabla_{\x}\cdot \widetilde{\bm \psi})\dd \x
+
\int_{\mathcal{X}_t}\widetilde{\bm \psi}^\top (\nabla_{\x}\rho(\x))\dd \x \\
=&\oint_{\partial \mathcal{X}_t} \rho(\x)(\widetilde{\bm \psi}\cdot \bm \nu)\dd S=0,
\end{align*}
because the boundary condition $\widetilde{\bm \psi}\cdot \bm \nu=0$ on $\partial \mathcal{X}_t$. 
So 
\begin{align*}
-&\int_{\mathcal{X}_t}\rho(\x) (\nabla_{\x}\cdot \widetilde{\bm \psi})\dd \x\\
=&\int_{\mathcal{X}_t}\widetilde{\bm \psi}^\top (\nabla_{\x}\rho(\x))\dd \x=\int_{\mathcal{X}_t}\nabla_{\x}\rho(\x)\cdot \widetilde{\bm \psi}\dd \x.
\end{align*}
Therefore, in $\mathcal{X}^t$, by performing integration by parts, we have 
\begin{equation*}
 \begin{aligned}
 & \frac{\dd}{\dd \epsilon} \Big|_{\epsilon = 0} \KL(\rho_{[\bm \phi^{\epsilon}]} || \rho^{*}) \\
 & = \int_{\mathcal{X}^t} - \rho_{[\bm \phi]} (\nabla_{\x} \cdot \widetilde{ \bm \psi}) + \rho \nabla V \cdot \widetilde{\bm \psi} \dd \x \\
 & = \int_{\mathcal{X}^t} (\nabla \rho + \rho \nabla V) \cdot \widetilde{\bm \psi} \dd \x, \\
 \end{aligned}
\end{equation*}
which implies that
\begin{equation}
 \frac{\delta \KL(\rho_{[\bm \phi]} || \rho^{*})}{\delta \bm \phi} = \nabla \rho + \rho \nabla V 
\end{equation}

Recall $V = - \ln \rho^{*}$. 
One can notice that if $F$ is an identity matrix, the result in \eqref{Var_phi} can be written as
\begin{equation*}
- \mathbb{E}_{\z \sim \rho_0} [\mathrm{trace} (\nabla_{\z} \bm{\psi} + \nabla \ln \rho^{*} \bm{\psi}^{\rm T})],
\end{equation*}
which is exactly the form given by the Stein operator in \cite{liu2016stein}.

\section{Proof of Theorem \ref{thm:converge}}
\begin{proof}
Let $\X \in \mathbb{R}^D$ be vectorized $\{ \x_i \}_{i = 1}^N$, that is
$$\X = (x_1^{(1)},\ldots x_N^{(1)}, \ldots x_1^{(d)} \ldots x_N^{(d)} ),$$
where $D = N \times d$.
Recall that $V(\x) = - \ln \rho^{*}$. 
For a sufficient smooth target distribution $\rho^{*}(\x)$, it is easy to show that 
\begin{equation*}
 \mathcal{F}_h ( \{ \x_i \}) = \frac{1}{N} \sum_{i=1}^N \left( \ln \left( \frac{1}{N} \sum_{j=1}^N K(\x_i, \x_j) \right) + V(\x_i) \right)
\end{equation*}
is continuous, coercive and bounded from below as a function of $\X \in \mathbb{R}^D$. We denote $\mathcal{F}_h ( \{ \x_i \})$ by $\mathcal{F}_h (\X)$.

For any given $\{ \x_i^n \}_{i=1}^N$, recall
\begin{equation*}
 J_n(\X) = \frac{1}{2 \tau} \|\X - \X^n\|^2 + \mathcal{F}_h(\X), 
\end{equation*}
where $\| \cdot\|^2_{\X}$ is a norm for $\X$, defined by
\begin{equation*}
\|\X - \X^n\|^2_{\X} = \frac{1}{N} \sum_{i=1}^N \| \x_i - \x_i^n \|^2.
\end{equation*}

Since
\begin{equation*}
\mathcal{S} = \{ J(\bm{\X}) \leq J(\bm{\X}^n) \}
\end{equation*}
is a non-empty, bounded, and closed set, by the coerciveness and continuity of $\mathcal{F}_h(\X)$, $J_n(\X)$ admits a global minimizer $\X^{n+1}$ in $\mathcal{S}$. Since $\X^{n+1}$ is a global minimizer of $J(\X)$, we have
\begin{equation*}
\frac{1}{2 \tau} \| \X^{n+1} - \X^{n} \|^2_{\X} + \mathcal{F}_h(\X^{n+1}) \leq \mathcal{F}_h(\X^{n}), 
\end{equation*}
which gives us equation \eqref{eq:decrease}.

For series $\{ \X^n \}$, since
\begin{equation*}
\| \X^{k} - \X^{k-1} \|^2_{\X} \leq 2 \tau (\mathcal{F}_h(\X^{k-1}) - \mathcal{F}_h(\X^{k})),
\end{equation*}
we have
\begin{equation*}
\sum_{k=1}^n \| \X^{k} - \X^{k-1} \|^2_{\X} \leq 2 \tau (\mathcal{F}_h(\X^{0}) - \mathcal{F}_h(\X^{n})) \leq C,
\end{equation*}
for some constant $C$ that is independent with $n$. Hence
\begin{equation*}
\lim_{n\rightarrow\infty} \| \X^{n} - \X^{n-1} \|_{\X} = 0,
\end{equation*}
which indicates the convergence of $\{ \X^n \}$. Moreover, since
\begin{equation*}
\X^{n} = \X^{n-1} - \tau \nabla_{\X} \mathcal{F}_h(\X^{n}),
\end{equation*}
we have
\begin{equation*}
\lim_{n \rightarrow \infty} \nabla_{\X} \mathcal{F}_h(\X^{n}) = 0,
\end{equation*}
so $\{\X^n\}$ converges to a stationary point of $\mathcal{F}_h(\X)$.
\end{proof}

\end{appendices}

\bibliographystyle{spmpsci}
\bibliography{VI}


%
%


\end{document}